\documentclass[11pt,a4paper]{article}

\usepackage[paper=a4paper,margin=1in]{geometry}%

\usepackage[utf8]{inputenc}
\usepackage{graphicx}
\graphicspath{ {./images/} }
\usepackage{xcolor}

\usepackage{hyperref}
\hypersetup{
    colorlinks=true,
    linkcolor=blue,
    filecolor=magenta,      
    urlcolor=cyan,
    citecolor=blue
}

\usepackage{todonotes}
\usepackage{url} 
\usepackage[acronym]{glossaries}

\usepackage{titlesec} 
\titlespacing\subsection{0pt}{5pt}{0pt} 
\usepackage[framemethod=default]{mdframed}

\newcommand{\artref}[2]{\hyperref[art:#2]{\textsf{\textbf{[#1]}}}}

\newcommand{\ArtA}{\artref{A}{A}}
\newcommand{\ArtB}{\artref{B}{B}}
\newcommand{\ArtC}{\artref{C}{C}}
\newcommand{\ArtD}{\artref{D}{D}}
\newcommand{\ArtE}{\artref{E}{E}}
\newcommand{\ArtF}{\artref{F}{F}}
\newcommand{\ArtG}{\artref{G}{G}}
\newcommand{\ArtH}{\artref{H}{H}}
\newcommand{\ArtI}{\artref{I}{I}}
\newcommand{\ArtJ}{\artref{J}{J}}
\newcommand{\ArtK}{\artref{K}{K}}
\newcommand{\ArtL}{\artref{L}{L}}
\newcommand{\ArtM}{\artref{M}{M}}
\newcommand{\ArtN}{\artref{N}{N}}
\newcommand{\ArtO}{\artref{O}{O}}
\newcommand{\ArtP}{\artref{P}{P}}
\newcommand{\ArtQ}{\artref{Q}{Q}}
\newcommand{\ArtR}{\artref{R}{R}}
\newcommand{\ArtS}{\artref{S}{S}}
\newcommand{\ArtT}{\artref{T}{T}}
\newcommand{\ArtU}{\artref{U}{U}}
\newcommand{\ArtV}{\artref{V}{V}}
\newcommand{\ArtW}{\artref{W}{W}}
\newcommand{\ArtX}{\artref{X}{X}}
\newcommand{\ArtY}{\artref{Y}{Y}}
\newcommand{\ArtZ}{\artref{Z}{Z}}
\newcommand{\ArtAA}{\artref{AA}{AA}}
\newcommand{\ArtAB}{\artref{BB}{BB}}
\newcommand{\ArtAC}{\artref{CC}{CC}}
\newcommand{\ArtAD}{\artref{DD}{DD}}
\newcommand{\ArtAE}{\artref{EE}{EE}}
\newcommand{\ArtAF}{\artref{FF}{FF}}
\newcommand{\ArtAG}{\artref{GG}{GG}}
\newcommand{\ArtAH}{\artref{HH}{HH}}

\usepackage{tcolorbox}

\newcounter{note}
\definecolor{noteHead}{RGB}{50,62,72}
\definecolor{noteMain}{RGB}{166,217,214}

\newenvironment{note}[1][]{
\begin{tcolorbox}[colback=noteMain, colframe=noteHead, title= \textsf{\large{\textbf{Note~\thenote. #1}}}]
\refstepcounter{note}
   \rmfamily
   \par\medskip}
{  \medskip\end{tcolorbox}}

\usepackage{tcolorbox}

\newcounter{example}
\definecolor{exHead}{RGB}{118,103,174}
\definecolor{exMain}{RGB}{142,209,241}

\newenvironment{example}[1][]{
\begin{tcolorbox}[colback=exMain, colframe=exHead, title= \textsf{\large{\textbf{Example~\theexample. #1}}}]
\refstepcounter{example}
   \rmfamily
   \par\medskip}
{  \medskip\end{tcolorbox}}

\usepackage{tcolorbox}

\newcounter{stage}
\newcommand{\stage}[1]{\section{Stage~\thestage.~#1\label{stg:\thestage}}\refstepcounter{stage}}

\newcommand{\stageref}[1]{\hyperref[stg:#1]{#1}}

\newcounter{ActivityCounter}
\refstepcounter{ActivityCounter}

\newcommand{\Activity}[2]{\subsection{Activity~\theActivityCounter: {#1}}\label{#2}\refstepcounter{ActivityCounter}}

\newcommand{\actref}[2]{\hyperref[#1]{#2}}

\makeglossaries

\title{AMLAS}
\author{colin.paterson }
\date{December 2020}

\begin{document}


\begin{titlepage}
	{\centering
 	{\Large{Guidance on the Assurance of Machine Learning in Autonomous Systems (AMLAS)
 	}}\par\vspace{1cm}
Richard Hawkins, Colin Paterson, Chiara Picardi, Yan Jia, Radu Calinescu and Ibrahim Habli. \par \vspace{0.5cm}
Assuring Autonomy International Programme (AAIP), University of York, UK

\texttt{firstname.lastname@york.ac.uk}

\textbf{Version 1, February 2021} 
\par\vspace{5mm}
}


\textbf{Abstract.} 
Machine Learning (ML) is now used in a range of systems with results that are reported to exceed, under certain conditions, human performance. Many of these systems, in domains such as healthcare , automotive and manufacturing, exhibit high degrees of autonomy and are safety critical. Establishing justified confidence in ML forms a core part of the safety case for these systems. In this document we introduce a methodology for the \textbf{A}ssurance of \textbf{M}achine \textbf{L}earning for use in \textbf{A}utonomous \textbf{S}ystems (AMLAS). AMLAS comprises a set of safety case patterns and a process for (1) systematically integrating safety assurance into the development of ML components and (2) for generating the evidence base for explicitly justifying the acceptable safety of these components when integrated into autonomous system applications.

\vspace{11cm}
\begin{centering}
	\textsf{The material in this document is provided as guidance only. No responsibility for loss occasioned to any person acting or refraining from action as a result of this material or any comments made can be accepted by the authors or The University of York.}
\end{centering}

\end{titlepage}

\tableofcontents
\newpage

\section{Introduction}
Machine Learning (ML) is now used in a range of systems with results that are reported to exceed, under certain conditions, human performance \cite{nagendran2020artificial}. Many of these systems, in domains such as healthcare \cite{topol2019high}, automotive \cite{koopman2017autonomous} and manufacturing \cite{jaradat2017challenges}, exhibit high degrees of autonomy and are safety critical \cite{burton2020a}. Establishing justified confidence in ML forms a core part of the safety case for these systems \cite{picardi2020assurance}. We introduce a methodology for the \textbf{A}ssurance of \textbf{M}achine \textbf{L}earning for use in \textbf{A}utonomous \textbf{S}ystems (AMLAS). AMLAS comprises a set of safety case patterns and a process for (1) systematically integrating safety assurance into the development of ML components and (2) for generating the evidence base for explicitly justifying the acceptable safety of these components when integrated into autonomous system applications.

AMLAS scope covers the following ML lifecycle stages: ML safety assurance scoping, safety requirements elicitation, data management, model learning, model verification and model deployment. In particular, the ML safety assurance scoping and the safety requirements elicitation stages explicitly establishes the fundamental link between the system-level hazard and risk analysis and the ML safety requirements. That is, AMLAS takes a whole system approach to ML assurance in which safety considerations are only meaningful once scoped within the wider system and operational context. The ML safety requirements are then used to weave the safety considerations into the ML stages in the subsequent phases. For each phase, we define a safety argument pattern that can be used to explain how and the extent to which the generated evidence supports the relevant ML safety claims, explicitly highlighting key assumptions, tradeoffs and uncertainties.

\section{Overview of AMLAS}

\begin{figure}[h]
    \centering
    \includegraphics[width=1\linewidth]{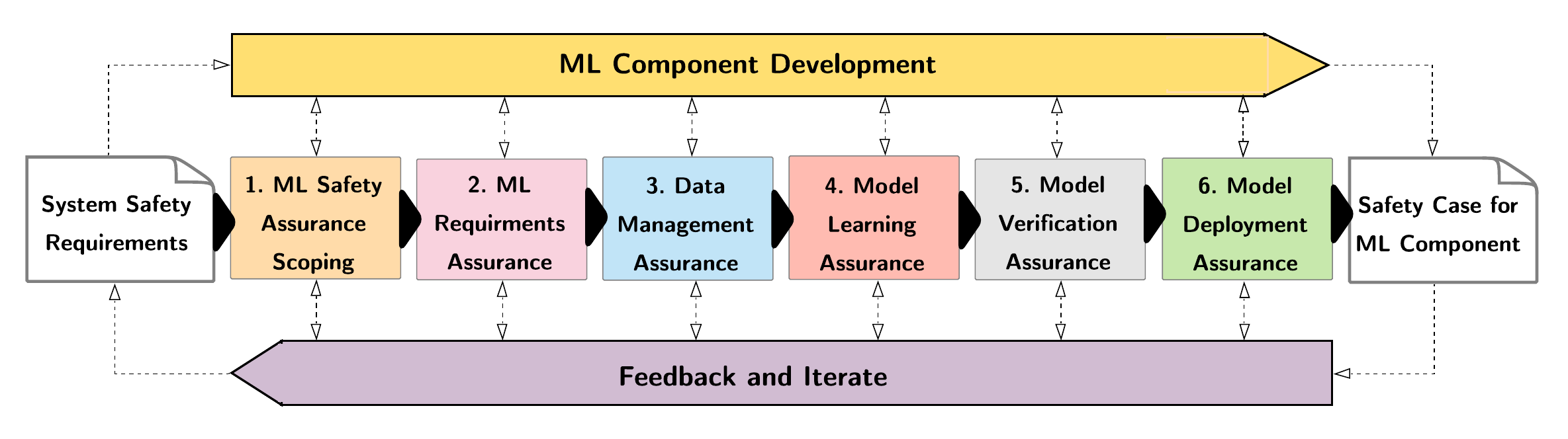}
    \caption{Overview of the AMLAS Process}
    \label{fig:OverviewOfAMLAS}
\end{figure}

Figure~\ref{fig:OverviewOfAMLAS} shows an overview of the six stages of the AMLAS process. For an ML component in a particular system context, the AMLAS process supports the development of an explicit safety case for the ML component. The AMLAS process requires as input the system safety requirements generated from the system safety process. The assurance activities are performed in parallel to the development process of the ML component. Further, the AMLAS process is iterative, as indicated by the feedback in Figure~\ref{fig:OverviewOfAMLAS}. Each stage of the AMLAS process is linked to the ‘Feedback and Iterate’ thread and could trigger the need to reconsider information generated or consumed by other stages. This is also necessary because of the interdependencies between the different stages, e.g. an activity in one stage might use artefacts produced by another activity in a previous stage. 

The stages of AMLAS may therefore be performed multiple times throughout the development of the ML component. For example, verification activities may reveal that ML safety requirements are not met by the ML component under some conditions. Depending upon the nature of the findings, this may require that stages such as model learning or data management must be revisited, or even that the ML requirements themselves must be reconsidered.

In this document, each AMLAS stage is structured as follows:
\begin{itemize}
    \item Objectives of the stage 
    \item Inputs to, and outputs of, the stage
    \item Description of the stage, including development and assurance activities and associated assurance artefacts and safety argument pattern
\end{itemize}

The description of each stage details the activities to be undertaken and the artefacts produced or required by the activities. The description also discusses common issues and misunderstandings relating to each activity; these are generally provided as notes or examples. Importantly, each stage concludes with an activity for instantiating a safety argument pattern based on the artefacts and evidence generated in the stage.

We adopt a commonly-used definition of a safety case as a ``structured argument, supported by a body of evidence that provides a compelling, comprehensible and valid case that a system is safe for a given application in a given operating environment.''~\cite{m2017a}. A safety case pattern documents a reusable argument structure and types of evidence that can be instantiated to create a specific safety case instance \cite{kelly1997safety}. 

\section{Using this Document}

The aim of this document is to provide guidance on how to systematically integrate safety assurance into the development of ML components. A primary outcome of this integration is an explicit and structured safety case. More specifically, AMLAS offers a set of argument patterns, and the underlying assurance activities, that can be instantiated in order to develop the ML safety cases. 

The scope of AMLAS is limited to the ML component. As such, this document should not be used in isolation from other standards and guidelines that specify best practices in safety-critical systems (e.g. ARP4754A \cite{arp4754a2010guidelines}),  domain-specific requirements (e.g. CONSORT-AI \cite{liu2020reporting} or ISO/PAS 21448 \cite{iso2019pas}) or safe autonomy considerations (e.g. UL4000 \cite{UL4600} or SCSC-153A \cite{SASWG2020}). For example, the system-level safety requirements, including acceptable risk targets, are a fundamental input to the AMLAS process. These requirements are expected to be generated by domain experts or derived from the relevant regulatory requirements.

AMLAS has a primary focus on off-line supervised learning. Off-line supervised learning, particularly applied to classification tasks, is currently the predominant application of ML for autonomous systems. Other types of ML such as reinforcement learning may also benefit from this guidance, particularly with regard to safety requirements and data management.

This document is aimed at
\begin{enumerate}
    \item safety engineers who are interested in determining the ML-specific safety considerations and evaluating the impact of the ML component on the system-level hazards and risks
    \item  ML developers who are interested in deriving and satisfying the safety requirements allocated to ML components
    \item other stakeholders who require assurance that the safety considerations have been explicitly and systematically considered
\end{enumerate}

The intended user of AMLAS is expected to have a basic understanding of machine learning, safety engineering and autonomous systems. Interested readers are encouraged to consult these practical and introductory resources:
\begin{itemize}
\item Autonomous systems: \cite{koopman2017autonomous,burton2020a}
\item Safety cases: ~\cite{kelly1997safety,group2018a}
\item Assurance practices: \cite{ashmore2019a, AAIP-BoK, UL4600, SASWG2020}
\item Machine learning: ~\cite{geron2017hands, franoischollet2017learning, sutton2018reinforcement, TensorFlow-Online, Keras-Online, lapan2018deep, lane2019natural}.
\end{itemize}

In AMLAS, the argument patterns are represented using the Goal Structuring Notation (GSN)~\cite{group2018a}. GSN is a graphical notation for explicitly capturing safety arguments that is widely used in many industries for documenting safety cases. For a detailed description of the notation, the reader is advised to consult the publicly available GSN standard~\cite{group2018a}.

Throughout the document, the use of "shall" indicates a required element of the guidance. Information marked as a ``NOTE'' or ``EXAMPLE'' is only used for clarification of the associated activities. A ``NOTE'' provides additional information, for clarification or advice purposes. An ``EXAMPLE'' is used to illustrate a particular point that is specific to a domain or technology. An example presented in this document is not meant to be exhaustive. Planned case studies and future experiments will provide more complete examples.

\clearpage
\stage{ML Safety Assurance Scoping}

\subsection*{Objectives}
\begin{enumerate}
\item Define the scope of the safety assurance process for the ML component.
\item Define the scope of the safety case for the ML component.
\item Create the top-level safety assurance claim and specify the relevant contextual information for the ML safety argument.
\end{enumerate}

\subsection*{Inputs to the Stage}
\begin{itemize}
\item[\ArtA]: System Safety Requirements
\item[\ArtB]: Description of Operating Environment of System 
\item[\ArtC]: System Description
\item[\ArtD]: ML Component Description
\item[\ArtF]: ML Assurance Scoping Argument Pattern
\end{itemize}

\subsection*{Outputs of the Stage}
\begin{itemize}
\item[\ArtE]: Safety Requirements Allocated to ML Component
\item[\ArtG]: ML Safety Assurance Scoping Argument
\end{itemize}

\subsection*{Description of the Stage}

\begin{figure}[h]
    \centering
    \includegraphics[width=0.7\linewidth]{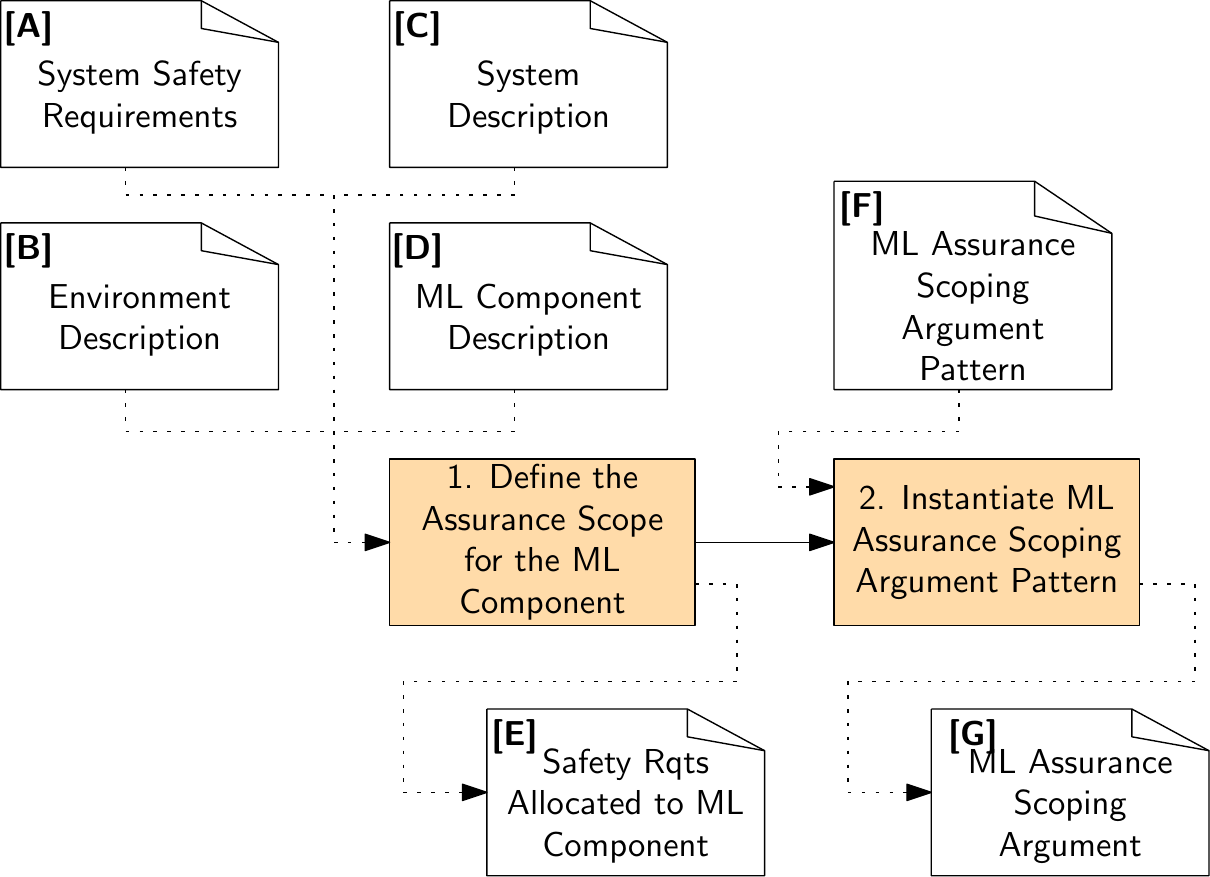}
    \caption{ AMLAS ML Assurance Scoping Process}
    \label{fig:ScopingProcess}
\end{figure}

As shown in Figure~\ref{fig:ScopingProcess}\footnote{In the AMLAS process diagrams, rectangles represent activities. Document symbols represent input or output artefacts. Each document symbol has a unique ID (top left) that is used to refer to the artefact in the guidance text or the argument pattern, e.g. \ArtA~ is a reference to artefact A.}, this stage consists of two activities that are performed to define the safety assurance scope for an ML component. The artefacts generated from this stage are used to instantiate the ML safety assurance scoping argument pattern as part of Activity~\ref{act:scopePattern}. An ML component comprises an ML model, e.g. a neural network, that is deployed onto the intended computing platform (i.e. comprising both hardware and software).  

Additional guidance on the use of ML for autonomous systems can be found at~\cite{AAIP-BoK}.

\Activity{Define the Safety Assurance Scope for the ML Component \label{art:E} \ArtE}{act:scope}
This activity requires as input the system safety requirements (\ArtA), descriptions of the system and the operating environment (\ArtB, \ArtC), and a description of the ML component that is being considered (\ArtD). These inputs shall be used to determine the safety requirements that are allocated to the ML component.

The safety requirements allocated to the ML component shall be defined to control the risk of the identified contributions of the ML component to system hazards.  This shall take account of the defined system architecture and the operating environment. At this stage the requirement is independent of any ML technology or metric but instead reflects the need for the component to perform safely with the system regardless of the technology later deployed. The safety requirements allocated to the ML component generated from this activity shall be explicitly documented (\ArtE). 

\begin{example}
Consider an autonomous driving application in which a subsystem may be required to identify pedestrians at a crossing. A component within the perception pipeline may have a requirement of the form ``When Ego is 50 metres from the crossing, the object detection component shall identify pedestrians that are on or close to the crossing in their correct position.''
\end{example}

\begin{note}
The allocation of safety requirements must consider architectural features such as redundancy when allocating the safety requirements to the ML component. Where redundancy is provided by other, non-machine-learnt components, this may reduce the assurance burden on the ML component that should be reflected in the allocated safety requirements.
\end{note}

\begin{note}
The contribution of the human as part of the broader system should be considered. A human may provide, for example, oversight or fallback in the case of failure of the ML component. These human contributions, and any associated human factors issues, e.g. automation bias~\cite{sujan-a}, should be reflected when allocating safety requirements to the ML component.
\end{note}

\subsection*{Artefact \ArtA: System Safety Requirements\label{art:A}}
\begin{figure}
    \centering
    \includegraphics[width=1\linewidth]{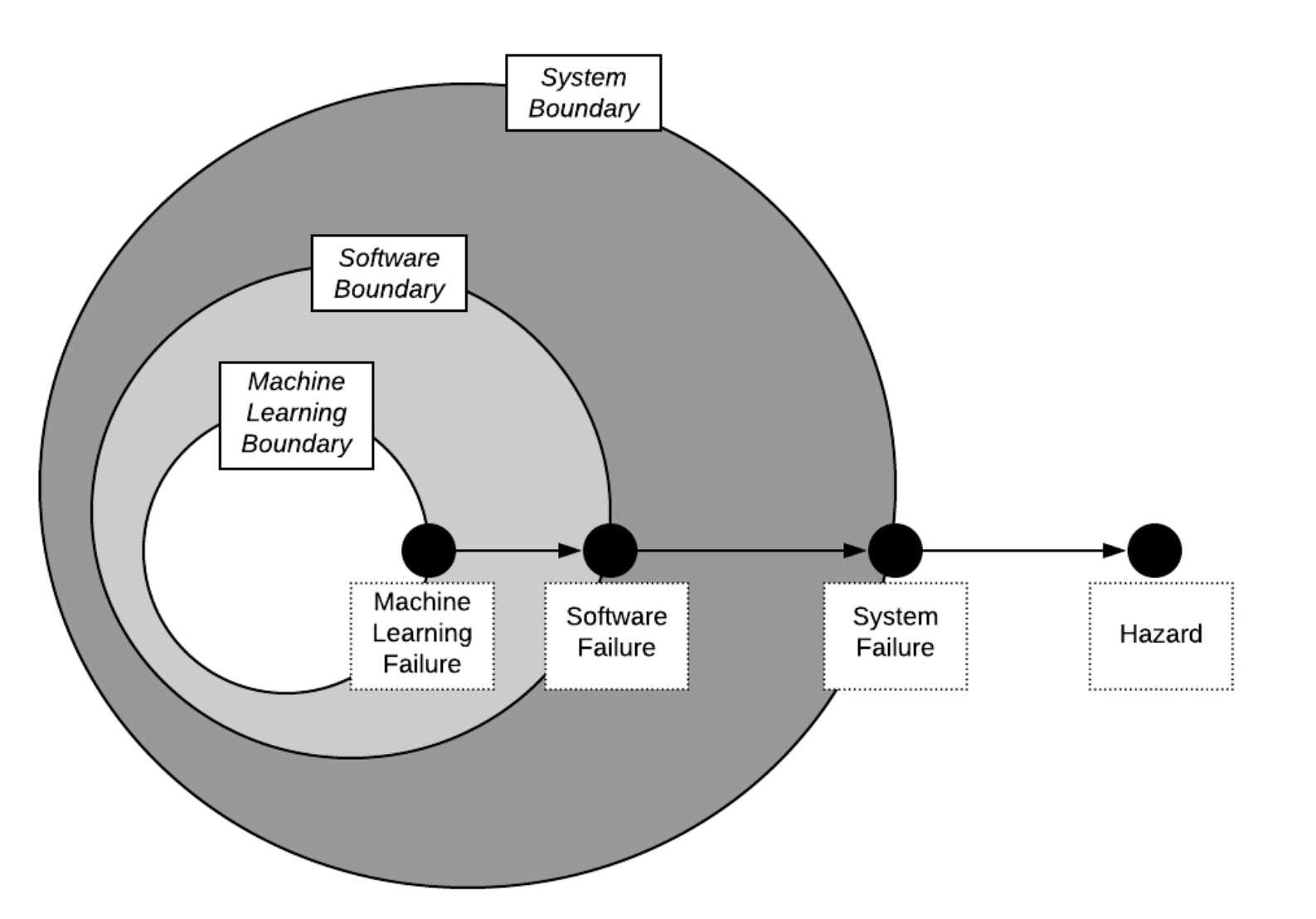}
    \caption{Simplified Chain of Failure Events (Adapted from~\cite{r2012a})}
    \label{fig:ChainFailureEvents}
\end{figure}

The safety requirements are generated from the system safety assessment process. Such a process covers hazard identification and risk analysis. Importantly, it shall determine the contribution, i.e. in the form of concrete failure conditions, that the output of the machine learning component makes to potential system hazards. A simplified linear chain of events that links a machine learning failure with a hazard is illustrated in Figure~\ref{fig:ChainFailureEvents}.

\begin{note}
It is important for the System Safety Requirements to explicitly capture risk acceptance criteria. Such criteria can generally be derived from the following sources:
\begin{itemize}
\item An existing system against which the proposed system can be compared. This may include a threshold of acceptance above (or below) current performance.
\item Existing standards for the acceptance of safety critical systems.
\item The views of stakeholders, including domain experts and users, who have a deep understanding of the context into which the system is to be deployed. These opinions may be founded on:
\begin{itemize}
    \item A scientific understanding of the processes at play e.g. vehicle dynamics.
    \item An understanding of the legal and ethical frameworks which govern the context.
    \item Personal experience.
    \item A study of similar systems and the lessons learnt from failures within these systems and operational contexts.
\end{itemize}
\end{itemize}
\end{note}

\subsection*{Artefact \ArtB: Description of System Environment\label{art:B}}
In determining the allocation of system safety requirements to the ML component it is crucial that the system environment is considered. The system environment considered during system safety requirement allocation shall be explicitly defined in Artefact \ArtB~ to ensure consistency when determining ML safety requirements.

\begin{example}	In developing an ML-based cancer screening system, the clinical pathway within which that  system is deployed will affect the safety requirements allocated to the ML system and must be documented.  The extent to which the output of the system is scrutinised by an independent clinician is also an important consideration that must be captured. \end{example}

\begin{example}	ML components for mortality predictions for patients admitted to the Intensive Care Unit (ICU) with COVID-19 are highly sensitive to criteria for ICU admission across hospitals, which in turn vary depending on ICU demand and capacity~\cite{futoma2020a}.\end{example}

\subsection*{Artefact \ArtC: System Description\label{art:C}}
The allocation of system safety requirements to the ML component shall also consider the system architecture. The system to which the ML component shall be deployed and the system architecture shall be explicitly defined in Artefact \ArtC. This helps to ensure it is correctly accounted for when determining ML safety requirements.

\begin{figure}[htb]
    \centering
    \includegraphics[width=1\linewidth]{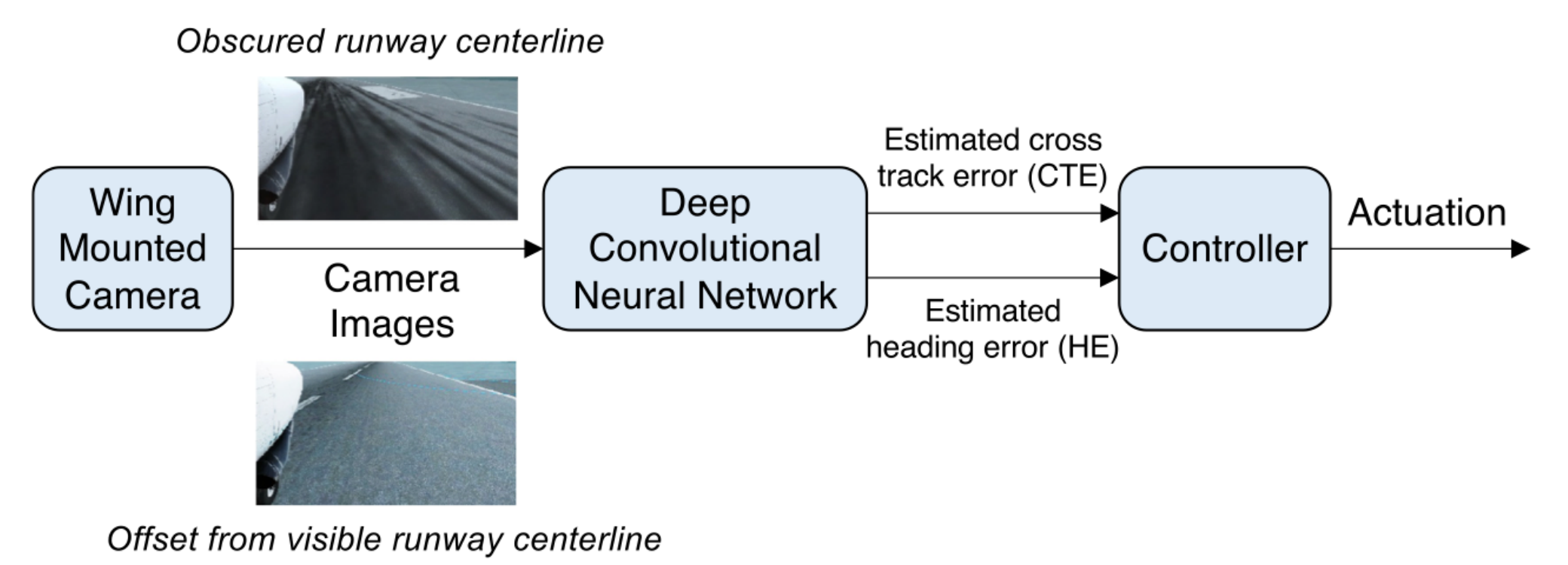}
    \caption{Example System Architecture (from~\cite{asaadi2020a})}
    \label{fig:ArtefactMLComponentdescription}
\end{figure}

\begin{note} The system description should include the inter-relationship between the ML component and other system components including sensors and actuators. Figure~\ref{fig:ArtefactMLComponentdescription} shows an example taken from~\cite{asaadi2020a} of the architecture for an autonomous taxiing capability in an unmanned aircraft system showing a Deep Convolutional Neural Network ML component in the context of other system components.
\end{note}

\subsection*{Artefact \ArtD: ML Component Description\label{art:D}}
This artefact describes the role and scope of the component within the system of which it is part, and the interfaces to which it is exposed.

\begin{example}	The primary input to an ML-based cancer screening system may come from different types of X-ray imaging devices or different configurations of the same device type.  The performance of different types and configurations of devices may vary so the allowable types and configurations must be defined. The output from the component is a medical diagnosis provided to a clinician.
\end{example}

\Activity{Instantiate ML Safety Assurance Scoping Argument Pattern \label{art:G} \ArtG}{act:scopePattern}
This activity requires as input the ML safety assurance scoping argument pattern (\ArtF), as well as the artefacts from Activity~\actref{act:scope}{2} (\ArtA, \ArtB, \ArtC, \ArtD~ and \ArtE). 
The activity uses these artefacts to create an instantiated ML assurance scoping argument (\ArtG) which documents the ML safety assurance scoping argument for the ML component and provides references to the contextual artefacts.

\subsection*{Artefact \ArtF: ML Safety Assurance Scoping Argument pattern}\label{art:F}

\begin{figure}[h]
    \centering
    \includegraphics[width=1\linewidth]{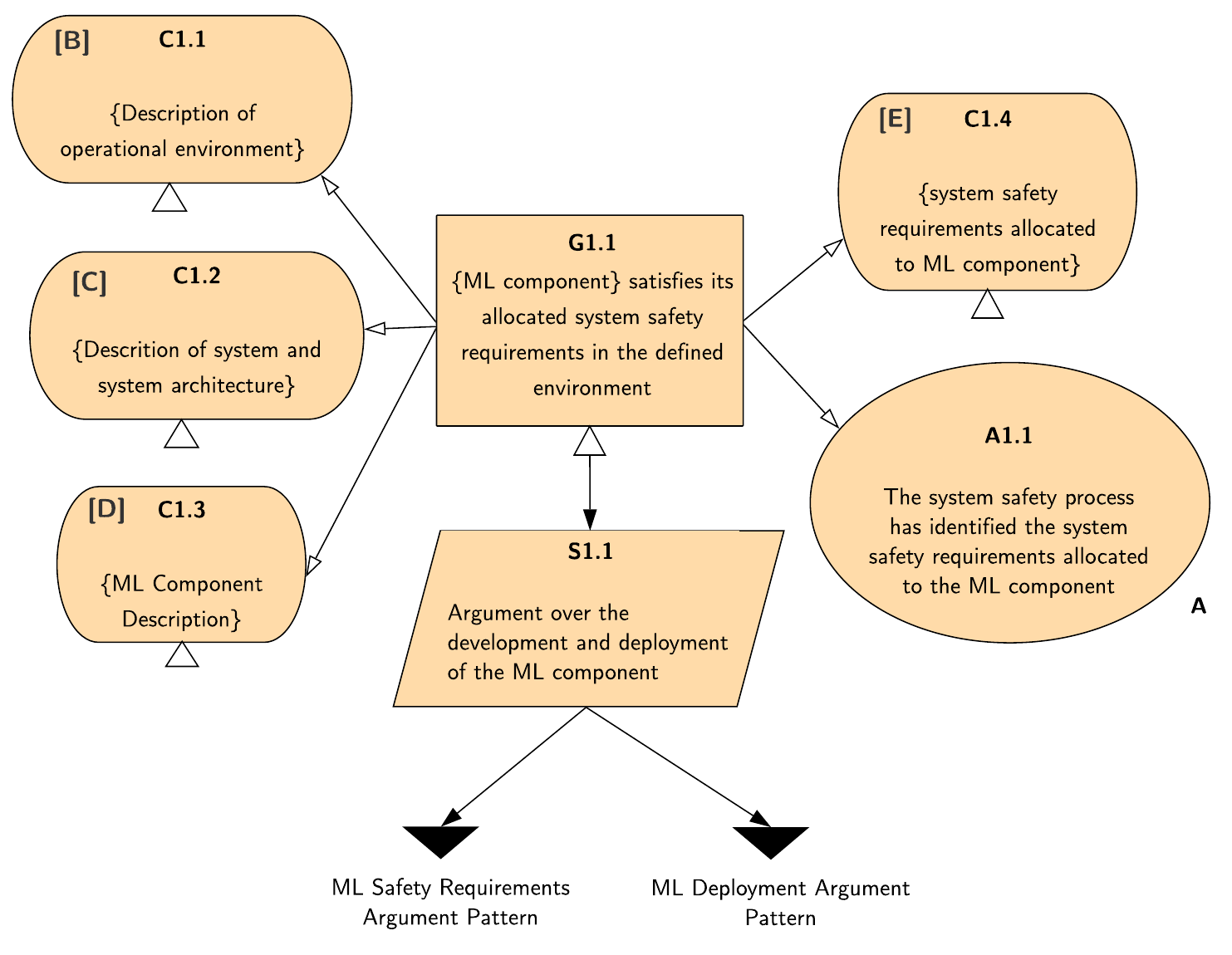}
    \caption{\ArtF~:~Argument Pattern for ML Safety Assurance Scoping} 
    \label{fig:ArgumentScoping}
\end{figure}

The argument pattern relating to this stage is shown in Figure~\ref{fig:ArgumentScoping} and key elements from the pattern are described in the following sections.

\subsection*{G1.1}
The top claim in this argument pattern represents the starting point for the safety argument for the ML component by claiming that the system safety requirements that have been allocated to the component are satisfied in the defined environment. As such, this claim provides the link to the higher level system safety argument of which it is a part. The safety claim for the ML component is made within the context of the information that was used to establish the safety requirements allocation including the descriptions of the system and software architectures (\ArtC) and operational environment (\ArtB), and the description of the ML component (\ArtD). The allocated system safety requirements (\ArtE) are also provided as context. It is important to be able to show that the allocated safety requirements have been correctly defined, however this is part of the system safety process and is therefore outside of the scope of the ML safety assurance argument. An assumption to this effect is therefore made explicitly in this argument in A1.1. It should be noted that to assure the validity of this assumption, a full argument and evidence regarding the system safety requirements should be provided in the safety case for the overall system. The primary aim of the ML Safety Assurance Scoping argument is to explain and justify the essential relationship between, on the one hand, the system-level safety requirements and associated hazards and risks, and on the other hand, the ML-specific safety requirements and associated ML performance and failure conditions (as detailed in Stage~\stageref{2}). 

\subsection*{S1.1}
The approach that is adopted to support the ML safety claim is to split the argument into two parts. Firstly the development of the ML component is considered. This argument begins through the development of the ML safety requirements argument as discussed in Stage~\stageref{2} of the process. Secondly the deployment of the ML component is addressed. The deployment argument is considered in Stage~\stageref{6} of the process.

The instantiated ML safety assurance scoping argument and references to artefacts shall be documented for the ML component (\ArtG). Along with the instantiated arguments resulting from the other stages of the AMLAS process, this will constitute the safety case for the ML component.

\clearpage
\stage{ML Safety Requirements Assurance}

\subsection*{Objectives}
\begin{enumerate}
\item Develop the machine learning safety requirements from the allocated system safety \newline requirements.
\item Validate the machine learning safety requirements against the allocated safety requirements, the system and software architecture and operational environment.
\item Create an assurance argument, based on the evidence generated by meeting the first two objectives, that provides a clear justification for the ML safety requirements. This should explicitly explain the tradeoffs, assumptions and uncertainties concerning both the safety requirements and the process by which they are developed and validated.
\end{enumerate}

\subsection*{Inputs to the Stage}
\begin{itemize}
\item[\ArtE]: Safety Requirements Allocated to ML Component
\item[\ArtI]: ML Safety Requirements Argument Pattern
\end{itemize}

\subsection*{Outputs of the Stage}
\begin{itemize}
\item[\ArtH]: ML Safety Requirements
\item[\ArtJ]: ML Safety Requirements Validation Results
\item[\ArtK]: ML Safety Requirements Argument
\end{itemize}

\subsection*{Description of the Stage}
\begin{figure}[h]
    \centering
    \includegraphics[width=0.7\linewidth]{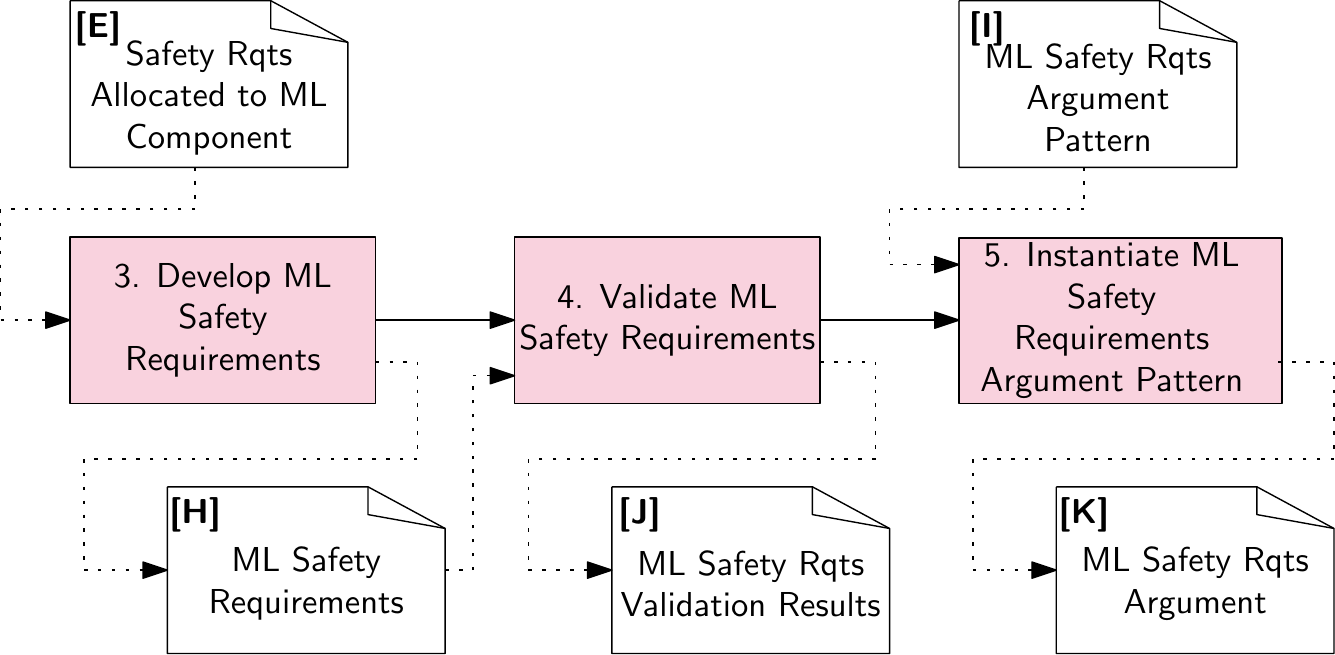}
    \caption{AMLAS ML Safety Requirements Assurance Process }
    \label{fig:ReqAssProcess}
\end{figure}

As shown in Figure~\ref{fig:ReqAssProcess}, this stage consists of three activities that are performed to provide assurance in the ML safety requirements. The artefacts generated from this stage are used to instantiate the ML safety requirements assurance argument pattern as part of Activity~\ref{act:InstMLSRA}. The scope of this stage is limited to the ML model, e.g. the mathematical representation of the neural network, that produces the intended output. 

\Activity{Develop ML Safety Requirements \label{art:H} \ArtH}{act:DevMLSR}
This activity requires as input the system safety requirements allocated to the ML component (\ArtE).

ML safety requirements shall be defined to control the risk of the identified  contributions of the ML component to system hazards, taking account of the defined system architecture and operating environment. This requires translating complex real world concepts and cognitive decisions into a format and a level of detail that is amenable to ML implementation and verification~\cite{rahimi2019a}. 

\begin{example}	In a system used for cancer screening based on X-ray images, a subset of the ML safety requirements will likely focus on defining an acceptable risk level by specifying true positive and false positive rates for diagnoses within the given environmental context~\cite{mckinney2020a}. This will take account of the performance achievable by human experts performing the same task, and the level of checking and oversight that is in place. \end{example}

\begin{example}From the safety requirement allocated to the ML component in Example 1 the concept of identifying a pedestrian at the system level must be translated into something meaningful for the ML model. Using knowledge derived from the system architecture, the ML safety requirement becomes  ``all bounding boxes produced shall be no more than 10\% larger in any dimension than the minimum sized box capable of including the entirety of the pedestrian''~\cite{gauerhof2020a}.\end{example}

\begin{note}	To some extent, the process of developing the machine learning safety requirements is similar to those of complex embedded software systems, e.g. in avionics systems or infusion pumps. However, due to the increasing transfer of complex perception and decision functions from an accountable human agent to the machine learning component, the difference between the implicit intentions on the component outputs and the explicit requirements that are used to develop, validate and verify the component is significant. This “semantic gap” exists in an open context for which a credible, let alone complete, set of concrete safety requirements is very hard to formalise~\cite{burton2020a}.\end{note}

\begin{note}In machine learning, requirements are often seen as implicitly encoded in the data. As such, under-specificity in requirements definition is an appealing feature. However, for rare events, such as safety incidents, this under-specificity poses a significant assurance challenge. The developers are still expected to assure the ability of the system to control the risk of these rare events based on concrete safety requirements against which the machine learning component is developed and tested.\end{note}

While there are likely to be a large range of requirements for the ML component, e.g. security, interpretability etc., the ML safety requirements should be limited to those requirements which impact the operational safety of the system.

\begin{note}	Other types of requirements such as security or usability should be defined as ML safety requirements only if the behaviours or constraints captured by these requirements influence the safety criticality of the ML output. 

‘Soft constraints’ such as interpretability may be crucial to the acceptance of an ML component especially where the system is part of a socio-technical solution. All such constraints defined as ML safety requirements must be clearly linked to safety outcomes.\end{note}

The ML safety requirements shall always include requirements for performance and robustness of the ML model. The requirements shall specifically relate to the ML outputs that the system safety assessment has identified as safety-related i.e. not just generic performance measures. 

\begin{note}
In this document, ML performance considers quantitative performance metrics, e.g. classification accuracy and mean squared error, whereas ML robustness considers the model’s ability to perform well when the inputs encountered are different \textit{but similar to} those present in the training data, covering both environmental uncertainty, e.g. flooded roads,
and system-level variability, e.g. sensor failure
\cite{ashmore2019a}.
\end{note}

\begin{note}
The performance of a model can only be assessed with respect to measurable features of the ML model. A model does not generally allow for us to measure risk or safety directly. Hence safety measures must be translated to relevant ML performance and robustness measures such as true positive count against a test set or point robustness to perturbations. Indeed not all misclassifications have the same impact on safety, e.g. misclassifying a speed sign of 40 mph as 30 mph is less impactful than misclassifying the same sign as 70 mph. 
\end{note}

\begin{note}	There is rarely a single performance measurement that can be considered in isolation for an ML component. For example for a classifier component, one may have to define a trade-off between false positives and false negatives. Over reliance on a single measure is likely to lead to systems which meet acceptance criteria but exhibit unintended behaviour~\cite{amodei2016a}. As such, the ML performance safety requirements should focus on reduction/elimination of sources of harm while recognising the need to maintain acceptable overall performance (without which the system, though safe, will not be fit for purpose). Performance requirements may also be driven by constraints on computational power, e.g. the number of objects which can be tracked. This is covered in more detail in Stage 6 (ML Deployment). 
\end{note}

\begin{note}	One useful approach to defining robustness requirements is to consider the dimensions of variation which exist in the input space. These may include, for example:
\begin{itemize}
\item variation within the domain, e.g. differences between patients of different ethnicity;
\item variation due to external factors, e.g. differences due to limitations of sensing technologies or effects of environmental phenomenon
\item variation based on a knowledge of the technologies used and their inherent failure modes.
\end{itemize}
\end{note}

\begin{example}
The ML safety requirement presented in Example 7 may now be refined into performance and robustness requirements~\cite{gauerhof2020a}. Example performance requirements may include:
\begin{itemize}
\item The ML component shall determine the position of the specified feature in each input frame within 5 pixels of actual position.
\item The ML component shall identify the presence of any person present in the defined area with an accuracy of at least 0.93
\end{itemize}
Example robustness requirements may include:
\begin{itemize}
\item The ML component shall perform as required in the defined range of lighting conditions experienced during operation of the system.
\item The ML component shall identify a person irrespective of their pose with respect to the camera.
\end{itemize}
\end{example}

Safety assessment shall not be limited to system-level activities. It is not a mere top-down process. Safety assessment shall be carried out in a continuous and iterative manner.  A detailed safety analysis of the outputs of the ML model shall be performed. This may identify new failure modes. The results of this analysis shall be fed back to the system-level safety assessment process for further examination such as reassessing the risk rating for a hazard. 

The activity of developing the ML safety requirements will likely identify implicit assumptions about the system or operating environment. Assumptions that are made shall be made explicit either as part of the description of the system environment or through defining additional safety requirements. Some domains refer to these as derived safety requirements.

\begin{note}	Derived safety requirement could relate to the assumed reliability and availability of sensor outputs or the specified thresholds of tolerable risk. In the case of the latter, current societal norms might accept the delegation of the interpretation of these often qualitative criteria to an accountable human, e.g. a quantified driver or a clinical professional.  Given the transfer of complex cognitive functions from human users to machine learning components, the process of developing concrete ML safety requirements will likely demand the interpretation of these thresholds at the design stage, typically through iterative interactions between domain experts and ML developers~\cite{habli2020a}.
\end{note}

The activity of developing the ML safety requirements may also identify emergent behaviour (potential behaviour of the ML component that could not be identified at the system level).  Where the emergent behaviour may contribute to a hazard, safety requirements shall be derived to ensure the emergent behaviour does not arise.

The ML safety requirements resulting from this activity shall be documented (\ArtH)

\Activity{Validate ML Safety Requirements \label{art:J}\ArtJ}{act:ValMLSR}
This activity requires as input the ML safety requirements (\ArtH) defined in Activity~\ref{act:DevMLSR}. The validity of these ML safety requirements shall be assured with respect to the intent of the allocated system safety requirements in the defined system and environmental context. 

The most commonly used approaches for validating ML safety requirements are:
\begin{itemize}
\item reviews: domain experts review all documentation to ensure that the specified ML safety requirements for the component will deliver the intended safe system operation.
\item simulation: in which a system which obeys the specified ML safety requirements for the component is constructed and the outcomes observed in the simulator for a set of operational scenarios. 
\end{itemize}

The results of the validation activities shall be documented (\ArtJ).

\Activity{Instantiate ML Safety Requirements Argument Pattern\label{art:K} \ArtK}{act:InstMLSRA}

This activity requires as input the ML safety requirements argument pattern (\ArtI), as well as the artefacts from the previous activities of this stage (\ArtE, \ArtH and \ArtJ ). The activity uses the activities and outputs form the previous stages to create an instantiated ML requirements argument (\ArtK).

\subsection*{Artefact \ArtI: ML Safety requirements argument pattern}\label{art:I}
\begin{figure}[htbp]
    \centering
    \includegraphics[width=1\linewidth]{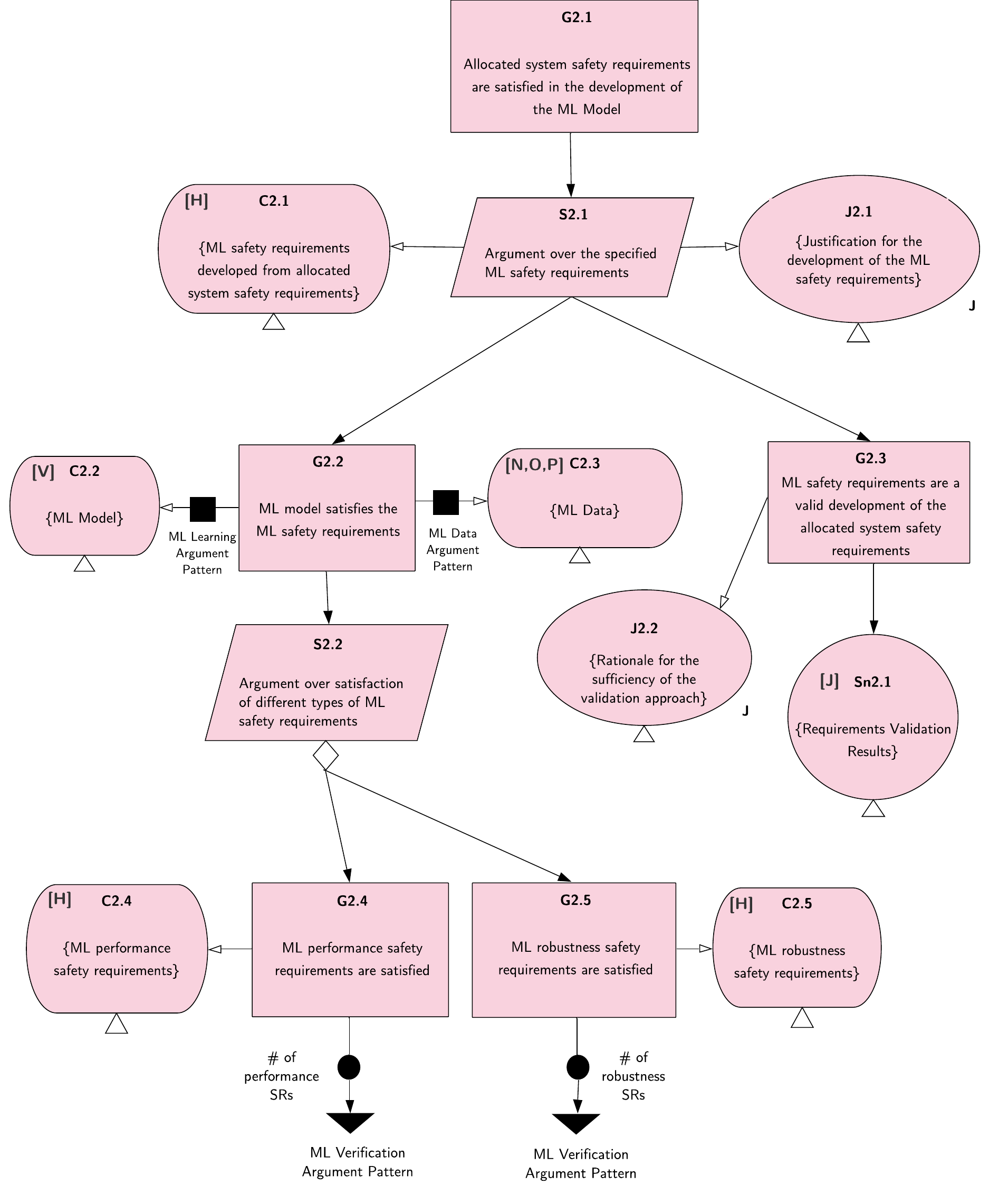}
    \caption{Assurance Argument Pattern for ML Safety Requirements  }
    \label{fig:PatternMLSR}
\end{figure}

The argument pattern relating to this stage of the AMLAS process is shown in Figure~\ref{fig:PatternMLSR} and the key elements are described in the following sections.

\subsection*{G2.1}
The top claim in this argument is that system safety requirements that have been allocated to the ML component (\ArtE) are satisfied by the model that is developed. This is demonstrated through considering explicit ML safety requirements defined for the ML model. 

\subsection*{S2.1}
The argument approach is a refinement strategy that justifies the translation of the allocated safety requirements into concrete ML safety requirements (\ArtH) as described in Activity~\ref{act:DevMLSR}.  Justification J2.1 is explicitly provided to explain the issues that were involved in translating the complex real world concepts and cognitive decisions into formats that are amenable to ML implementation. This should also explain and justify the scope of the ML safety requirements and whether any of the allocated system safety requirements were not fully specified as part of the ML safety requirements. Any such allocated requirements must be addressed as part of the system safety process. For example, allocated system safety requirements with real-time targets, which require the consideration of the performance of the underlying hardware, cannot be fully specified and tested merely by the ML model. As such these can only be meaningfully considered by also testing the integrated ML component (i.e. Stage 5). To support this strategy two subclaims are provided in the argument, one demonstrating that  the ML safety requirements are valid, and one concerning the satisfaction of those requirements. 

\subsection*{G2.3}
The validity claim is provided to demonstrate that the ML safety requirements are a valid development of the allocated system safety requirements. Evidence from the validation results (\ArtJ) obtained in Activity~\ref{act:ValMLSR} is used to support the validity claim. Justification J2.2 provides rationale for the validation strategy that was adopted for Activity~\ref{act:ValMLSR}.

\subsection*{G2.2}
This claim focuses exclusively on the ML safety requirements. The claim states that the ML safety requirements are satisfied by the ML model. The claim is made in the context of the ML model (\ArtV) that is generated and the data (\ArtN, \ArtO and \ArtP) that is used to create the model. Although the satisfaction of the ML safety requirements is demonstrated through verification evidence, it is also important, as for more traditional software, to provide assurance regarding the processes used for development. The ML Learning Argument Pattern (\ArtW) and the ML Data Argument Pattern (\ArtR) are therefore used to provide argument and evidence that the model (and learning process) and the data (and data management process) are sufficient and are discussed in detail in Stages~\stageref{4} and~\stageref{3} respectively. The link with assurance in these stages is established using Assurance Claim Points (ACPs) \cite{hawkins2011new} (indicated by the black squares). These represent points in the argument at which further assurance is required, focusing specifically here on how confidence in data management and model learning can be demonstrated. These ACPs can be supported through instantiation of the ML Data Argument Pattern (\ArtW) and the ML Data Argument Pattern (\ArtR) respectively.

\subsection*{S2.2}
This is a decomposition strategy based on the different types of ML safety requirements. As shown in Figure \ref{fig:PatternMLSR}, this will include claims regarding performance and robustness requirements, but may also include other types of ML requirements such as interpretability where these requirements are relevant to the system safety requirements. This is indicated by the ‘to be developed’ symbol, i.e. diamond, under the strategy.  

\subsection*{G2.4}
This claim focuses on the ML safety requirements that consider ML performance with respect to safety-related outputs. The defined ML safety requirements that relate to performance are provided as context to the claim. The argument considers each of these requirements in turn and provides a claim regarding the satisfaction of each requirement (G5.1 in the ML verification argument pattern \ArtAB). The satisfaction of each requirement will be demonstrated through verification activities. These are discussed in more detail in Stage~\stageref{5}.

\subsection*{G2.5}
This claim focuses on, and is stated in the context of, the ML safety requirements that consider ML robustness with respect to safety-related outputs. The defined ML safety requirements that relate to robustness are provided as context to the claim. The argument considers each of these requirements in turn and provides a claim regarding the satisfaction of each requirement (G5.1 in the ML verification argument pattern \ArtAB). The satisfaction of each requirement will be demonstrated through verification activities. These are discussed in more detail in Stage~\stageref{5}

\clearpage
\stage{Data Management}

\subsection*{Objectives}
\begin{enumerate}
\item Develop data requirements which are sufficient to allow for the ML safety requirements to be encoded as features against which the data sets to be produced in this stage may be assessed. 
\item Generate data sets in accordance with the data requirements for use in the development and verification stages, providing a rationale for those activities undertaken with respect to the ML safety requirements.
\item Analyse the data sets obtained by objective 2 to determine their sufficiency in meeting the data requirements. 
\item Create an assurance argument, based on the evidence generated by meeting the first three objectives, that provides a clear justification of the ML Data requirements. This should explicitly state the assumptions and tradeoffs made and any uncertainties concerning the data requirements and the processes by which they were developed and validated.
\end{enumerate}

\subsection*{Inputs to the Stage}
\begin{itemize}
    \item[\ArtH]: ML safety requirements
    \item[\ArtR]: ML data argument pattern
\end{itemize}

\subsection*{Outputs of the Stage}
\begin{itemize}
    \item[\ArtL]: Data requirements
    \item[\ArtM]: Data requirements justification report
    \item[\ArtN]: Development data
    \item[\ArtO]: Internal test data
    \item[\ArtP]: Verification data
    \item[\ArtQ]: Data generation log
    \item[\ArtS]: ML data validation results
    \item[\ArtT]: ML data argument
\end{itemize}

\subsection*{Description of the Stage}
\begin{figure}[htbp]
    \centering
    \includegraphics[width=0.9\linewidth]{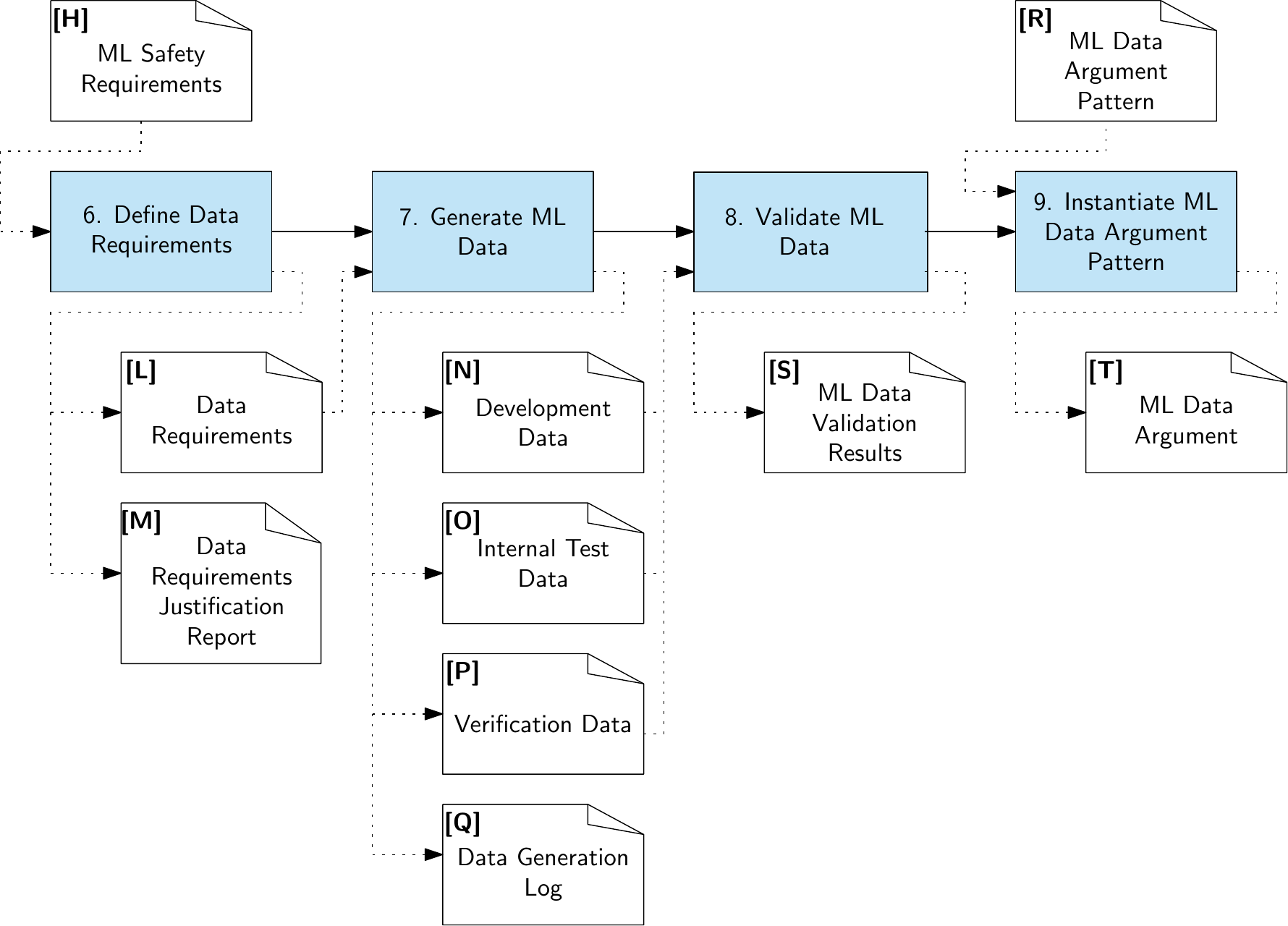}
    \caption{AMLAS ML Data Requirements Assurance Process }
    \label{fig:MLDRProcess}
\end{figure}

As shown in Figure~\ref{fig:MLDRProcess} , this stage consists of four activities that are performed to provide assurance in the ML data. The artefacts generated from this stage are used to instantiate the ML data  assurance argument pattern as part of Activity~\ref{act:instMLArgPat}.

Additional guidance on this stage can be found at~\cite{AAIP-BoK-Data}.

\Activity{Define Data Requirements}{act:defDR}
Data plays a particularly important role in machine learning with data encoding the requirements which will be embodied in the resulting ML model. ML data requirements shall therefore be defined to ensure it is possible to develop a machine learnt model that satisfies the ML Safety Requirements. This activity requires as input the ML safety requirements (\ArtH) as described in Stage~\stageref{2} and, from these requirements, data requirements (\ArtL) shall be generated. Of particular interest in the development of data requirements are those safety requirements which pertain to the description of the system environment.

\subsection*{Artefact \ArtL : Data Requirements\label{art:L}}
The ML data requirements shall specify the characteristics that the data collected must have in order to ensure that a model meeting the ML safety requirements may be created. ML data requirements shall include consideration of the relevance, completeness, accuracy and balance of the data~\cite{ashmore2019a}. These requirements shall explicitly state the assumptions made with respect to the operating environment and the data features required to encode the domain.

\begin{note}
ML data requirements will often focus on  specifying data which is necessary to ensure the robustness of the model in the context of the operational domain. This should relate to the dimensions of variation anticipated in the operational domain as enumerated in the ML safety requirements. 
\end{note}

ML data requirements relating to relevance shall specify the extent to which the data must match the intended operating domain into which the model is to be deployed. 

\begin{example}	For an ML component used for object detection on a vehicle the following may be defined as an ML data requirement for relevance: “Each data sample shall assume sensor positioning which is representative of that to be used on the vehicle”. This requirement is defined to ensure that images that provide a very low or very high viewpoint of the road (such as an aerial view) are not used in development.\end{example}

\begin{example}	For an ML component used for medical diagnosis based on X-Ray Images for use in UK Hospitals the following may be defined as an ML data requirement for relevance: “Each data sample shall be representative of those images gathered for Machines of type A, B, C which are in use in UK hospitals”. This requirement ensures that image artefacts due to image processing in machines used outside the UK are not used in development.\end{example}

ML data requirements relating to completeness shall specify the extent to which the development data must be complete with respect to a set of measurable dimensions of the operating domain. This can be done through reference to the anticipated dimensions of variation stated in the ML safety requirements (\ArtH) or defined by the operating context (\ArtB).

\begin{example}	The operational domain for an autonomous vehicle indicates that the vehicle is to operate at all times of day and that the ML component should be robust to changing light levels. An ML data requirement for completeness may state: “Data samples should be gathered at all times of day and under the following light conditions: sunlight, cloud, rural with headlights and urban street lighting”.\end{example}

\begin{example}	When building a model to determine the life expectancy for patients suffering from liver failure a MELD Score is commonly used which is calculated from four lab tests on the patient~\cite{meld-score}. Normal ranges for each of these results are known from historic data. A  completeness requirement may state that: “Data samples should as a minimum include patients with Bilirubin levels across the range of [5.13, 32.49]”.\end{example}

ML data requirements shall include requirements that specify the required accuracy of the development data. 

\begin{note}	Requirements may relate to the labelling of data samples. Label quality has a big impact on the reliability of the risk acceptance criteria. Deciding on these criteria involves subjective judgement and is prone to systematic and random errors~\cite{chen2019a}. In a study reported by Krause et al. the same ML model had a 30\% relative reduction in errors after switching from labels established by a majority vote of three retinal specialists to labels established by adjudication from the same specialists~\cite{krause2018a}. 
\end{note}

\begin{example}	Consider an ML safety requirement that all pedestrians should be identified within 50cm of their true position. Given that the pedestrians are not point masses but instead represented as coloured pixels in the image, an accuracy requirement must clearly specify the required position of the label including the positioning of labels for partially occluded objects. An example accuracy requirement may state that: “When labelling data samples, the position of all pedestrians shall be recorded as their extremity closest to the roadway”.\end{example}

ML data requirements relating to balance shall specify the required distribution of samples in the data sets. 

\begin{note}	Consider a classifier which is designed to identify one of n classes. A data set which is balanced with respect to the classes would present with the same number of samples for each class. More generally however balance may be considered with respect to certain features of interest, e.g. environmental conditions, gender, race etc. This means that a data set which is balanced with respect to the classes may present as biased when considering critical features of the data. \end{note}
	
\begin{example}	DeepMind’s ML model for detecting acute Kidney failure reports incredible accuracy and predictive power. However analysis~\cite{talby2019a} shows that the data used to train the model was overwhelmingly from male patients (93.6\%). In this case an ML data requirement for balance in the gender of the data sources should have been explicitly specified since this feature is relevant to the operating context of the model (which will be used for both male and female patients). Similarly, the data was collected from a set of individuals that lacked other forms of diversity. This could lead to the results in operation falling far short of those promised for the affected groups of patients.
\end{example}

\subsection*{Artefact \ArtM : Data Requirements Justification Report\label{art:M}}
A justification shall be provided that the specified ML data requirements are sufficient to ensure it is possible to develop a machine learnt model that satisfies the ML Safety Requirements. This justification shall be documented in a data requirements justification report (\ArtM). This will typically require an analysis of the data requirements to ensure that the intent of the ML safety requirements are maintained by the data to be collected. This may involve Expert review and statistical analysis techniques.

\Activity{Generate ML Data \label{art:N} \label{art:O}  \label{art:P}\ArtN,~\ArtO,~\ArtP}{act:GenMLD}
 Data shall be generated that meets the ML data requirements established in Activity~\ref{act:defDR}. This shall include three separate datasets: Development data \ArtN, Internal test data \ArtO~ and Verification data \ArtP\footnote{We use the term development data to include training and validation data as it is normally referred to in the ML literature. Development data is used to create a model which is then tested by the development team using the internal test data. Once a model is deemed fit for release by the development team only then is it exposed to the Verification data.}. The first two of these sets are for use in the development process (Stage 3) whilst verification set is used in model verification (Stage 4).

The generation of ML data will typically consider three sub-process: collection, preprocessing and augmentation.

\begin{em}Data collection\end{em} shall be undertaken to obtain data from sources that are available to the data collection team which sufficiently addresses the ML data requirements. This may involve reusing existing data sets where they are deemed appropriate for the context, or the collection of data from primary sources.

\begin{note}	It may be necessary to collect data from systems which are close to, but not identical to, the envisioned system. Such compromises and restrictions should be stated explicitly in the data generation log with a justification of why the data collected is still valid. \end{note}

\begin{example}	A vehicle gathering video data is an experimental variant of the proposed vehicle where variation in vehicle dynamics are assumed to have no impact on the video data with respect to prediction of distance to leading vehicles.\end{example}

\begin{note}	Where existing data sources are used, a rationale should be provided in the data generation log as to how these data sources may be transferred to the current domain and any assumptions concerning relevance should be stated explicitly. \end{note}

\begin{note} 	Where it is impossible to gather real world samples it is common to use simulators. These may be software or hardware in the loop. Where data is collected for such simulators the configuration data should be recorded to allow for repeatable data collection and to support systematic verification and validation of the simulator within the operational context. Such simulators might need to be subjected to a separate assurance or approval process, such as discussed in \cite{sargent2010verification} or similar to tool qualification in the aerospace guidance DO178C~\cite{r2012a}. \end{note}

\begin{em}Data Preprocessing\end{em} may be undertaken to transform the collected data samples into data that can be consumed by the learning process. This may involve the addition of labels, normalisation of data, the removal of noise or the management of missing features.

\begin{note}	Preprocessing of data is common and is not necessarily used to compensate for failures in the data collection process. Indeed the normalisation of data is often used to improve the performance of trained models. Consider for example pixel values in an image in the range [0,255]  scaled to be floating point values in the range [0,1]. \end{note}

\begin{example}	Handling missing data is particularly important when using clinical health data with missing data rates reported from 30\% to 80\%~\cite{hu2017a}.  The strategies used to tackle missing data should be stated explicitly in the data generation log with a clear rationale as to why the approach used is commensurate with the system under consideration.
When an ML component is used to predict cardiovascular disease many records are found to be missing blood pressure measurements. A preprocessing rationale may be : “Analysis of the data has shown a correlation between the recorded values for patient’s blood pressure and age. Linear regression on the training set is therefore used to impute blood pressure where it is not recorded”. \end{example}

\begin{note}
A common preprocessing activity is the addition of labels to data. This is particularly important in supervised learning where the labels provide a baseline, or ground truth, against which learnt models can be assessed. Whilst labelling may be trivial in some contexts this may not always be the case. For example, labelling may require a consensus of opinion for use in medical prognosis. In such cases a process to ensure consistent labelling should be developed, documented and enacted.  
\end{note}

\begin{example}	A set of images from retinal scans may be examined by clinical professionals to provide labels which reflect the appropriate diagnosis. A more advanced labelling process may be necessary if the system is required to identify regions of the image which are to be referred to experts. In this case labelling requires a region of the image to be specified as a closed boundary around those pixels in the image which relate to the region of concern.\end{example}

\begin{em}Data Augmentation\end{em} shall be undertaken to allow for the addition of data where it is infeasible to gather sufficient samples from the real world. This may occur when the real world system does not yet exist or where collecting such data would be too dangerous or prohibitively expensive. In such cases the data sets shall be augmented with data which is either derived from existing samples or collected from systems which act as a proxy for the real world.

\begin{example}
The field of computer vision utilises sophisticated models of environmental conditions exist~\cite{zhang2017towards} and, by collecting an image of an object under one controlled lighting condition, it is possible to augment the data set with examples of that one object under many simulated lighting conditions. The data generation log should document this augmentation process and provide a justification that the simulated conditions are a sufficient reflection of reality. \end{example}

\begin{example}
An ML classifier for cancer accepts chest X-RAYs of patients. The ML safety requirement states that the classifier should be robust to rotations in the image up to 15 degrees. Each sample in the collected data set may be rotated in 1 degree increments and labelled with the original images labels, appropriately translated.
\end{example}

Verification data is gathered with the aim of testing the models to breaking point. This requires a different mindset for the team engaged with collecting data for verification who are focused not on creating a model but finding realistic ways in which the model may fail when used in an operational system. Furthermore the nature of ML is that any single sample may be encoded into the training set and a specific model found which is able to avoid the failure associated with the sample. This does not mean that the resultant model is robust to a more general class of failure to which the sample belongs. It is imperative therefore that the information concerning verification data is hidden from the developers to ensure the models generated are robust to the whole class of failures and not just specific examples present in the verification data.

\begin{note}
The dimensions of variation are not independent and as such combinations of difficult situations are less likely to be included in a data set, used for development, which aims to represent normal operating behaviour. For example a vehicle is unlikely to be using high beam in foggy conditions on a rainy day where ice is present on the road and a vehicle is approaching on the incorrect side of the carriageway. As such this case, although within the operating domain of the vehicle, is unlikely to be found in the development set.
A good verification data set should particularly focus on challenging conditions which are within the operating domain and therefore such a case may be present in the verification data set.  
\end{note}

\subsection*{Artefact \ArtQ :Data Generation Log\label{art:Q}} 
It is possible for many data sets to be generated which meet the data requirements. Decisions made when collecting, processing and augmenting the data should therefore be recorded in order to explain how the data sets meet the data requirements. A data generation log (\ArtQ) shall be kept which details the decisions made in each sub-process to obtain data with the desired features.

\Activity{Validate ML Data \label{art:S}\ArtS}{act:ValMLD}
The ML data validation activity shall check that the three generated data sets are sufficient to meet the ML data requirements. The results of the data validation activity shall be explicitly documented (\ArtS). Data validation shall consider the relevance, completeness, and balance of the data sets.

Discrepancies identified between the data generated and the ML data requirement shall be justified. These justifications shall be captured as part of the data validation results (\ArtS).

\begin{note} 	Both financial and practical concerns can lead to data sets which are not ideal and, in such cases, clear rationale shall be provided. For example a young child crossing in front of a fast moving car may be a safety concern but gathering data for such events is not practicable. \end{note}

\begin{example}
The JAAD open dataset~\cite{rasouli2017a} is used as development data for an ML component used to detect pedestrians. The cost of gathering and processing data for road crossings is expensive and substantial effort has been undertaken to generate the JAAD dataset. The labelling of pedestrians and range of poses observed is extensive and is clearly relevant for a perception pipeline concerned with the identification of pedestrians. The range of crossings types observed is limited however and a justification may be required as to why this is relevant for the intended deployment. 
\end{example}

Validation of data relevance shall consider the gap between the samples obtained and the real world environment in which the system is to be deployed. Validation shall consider each of the sub-activities undertaken in data generation and provide a clear rationale for their use.

\begin{note}
	Any simulation used for data augmentation is necessarily a simplification of the real world with assumptions underpinning the models used. Validating relevance therefore requires the gaps between simulation and modelling to be identified and a demonstration that these gaps are not material in the construction of a safe system.
\end{note}

\begin{note}
 Validation should demonstrate that context-specific features defined in the ML safety requirements are present in the collected datasets. For example, for a pedestrian detection system for deployment on European roads the images collected should include road furniture of types that would be found in the anticipated countries of deployment.
\end{note}

\begin{example}
	Data gathered in US hospitals used for a UK prognosis system should state how local demographics, policies and equipment vary between countries and the impact of such variance on data validity. 
\end{example}

\begin{example}
    When data is collected using controlled trials, e.g. for medical imaging, a decision may be made to collect samples using a machine setup away from the hospital using non-medical staff. The samples may only be considered relevant if an argument can be made that the environmental conditions do not impact the samples obtained and that the experience of the staff has no effect on the samples collected. 
\end{example}	

Validation of data completeness shall demonstrate that the collected data covers all the dimensions of variation stated in the ML safety requirements sufficiently.  Given the combinatorial nature of input features validation shall seek to systematically identify areas which are not covered.

\begin{note}
 As the number of dimensions of variability and the granularity with which these dimensions are encoded increases, so the space that must be validated increases, combinatorially.  
\end{note}	

\begin{note}
For continuous variables the number of possible values is infinite. One possible approach is to use quantisation to map the continuous variables to a discrete space which may be more readily assessed.  Where quantisation is employed it should be accompanied by an argument concerning the levels used.
\end{note}

\begin{example}
Consider a system to identify road signs into 43 separate classes. Dimensions of variability are: Weather, time of day and levels of partial occlusion up to 70\%. Let us assume that we have categorised each dimension as:  
\begin{itemize}
\item Time : {early morning, mid morning, noon, late afternoon, evening, late evening, night}
\item Weather : {clear, rain light, rain heavy, fog light, fog heavy, snow light, snow heavy}
\item Occlusion (\%) : (0, 10, 20, 30, 40,50, 60, 70)
\end{itemize}
Validation may show that there are samples for each of the 43 * 7 x 7x 8 =16856 possible combinations. A systematic validation process will identify that the data sets are missing  e.g. no samples containing a 40mph sign in light rain with 50\% occlusion early in the morning.  Although for most practical systems completeness is not possible this process should provide evidence of those areas which are incomplete and why this is not problematic for assuring the resultant system.
\end{example}	

Validation of \textbf{\emph{data balance}} shall consider the distribution of samples in the data set. It is easiest to consider balance from a supervised classification perspective where the number of samples associated with each class is a key consideration. 

\begin{note}
At the class level assessing balance may be a simple case of counting the number of samples in each class, this approach becomes more complex however when considering the dimension of variation where specific combinations are relatively rare. More generally data validation shall include statements regarding class balance and feature balance supervised learning tasks.

\end{note}	

\begin{note}
Certain classes may naturally be less common and, whilst techniques such as data augmentation may help,  it may be difficult, or even impossible to obtain a truly balanced set of classes. In such cases, the imbalance shall be noted and a justification provided as part of the validation results to support the use of imbalance data in the operational context. 

\end{note}	

\begin{example}
Using the previous example we may count the number of samples at each level of occlusion to ensure that each level is appropriately represented in the data sets..

\end{example}	

Validation of \textbf{\emph{data accuracy}} shall consider the extent to which the data samples, and meta data added to the set during preprocessing (e.g. labels), are representation of the ground truth associated with samples. Evidence supporting the accuracy of data may be gathered through a combination of the following: 
\begin{itemize}
\item An analysis of the processes undertaken to collect data: e.g. A bush fire detection system using satellite imagery could ensure that at least 3 users have agreed on the label for each sample.
\item Checking subsets of samples by expert users: e.g. Where MRI images are generated with augmentation to simulate varying patient orientation within the scanner field an expert clinician will review a random sample of the resulting images to ensure that that they remain credible.
\item Ensuring diversity of data sources to avoid systematic errors in the data sets: e.g. Data for use in an earthquake detection system should make use of multiple sensors and locations such that sensor drift or atmospheric effects may be identified.
\end{itemize}

Where existing data sets are re-used, e.g. the JAAD pedestrian data set~\cite{rasouli2017a}, documentation concerning the process may be available. Even under these conditions additional validation tasks may be required to ensure that the labels are sufficient for the context into which the model is to be deployed. 

\Activity{Instantiate ML Data Argument Pattern \label{art:T} \ArtT}{act:instMLArgPat}

This activity requires as input the ML data argument pattern (\ArtR), as well as the artefacts from the previous activities of this stage (\ArtL, \ArtM, \ArtN, \ArtO, \ArtP, \ArtQ and \ArtS). The activity uses the activities and outputs from the previous stages to create an instantiated ML data argument (\ArtT).

\subsection*{Artefact \ArtR: ML Data argument pattern}\label{art:R}
\begin{figure}[htbp]
    \centering
    \includegraphics[width=1\linewidth]{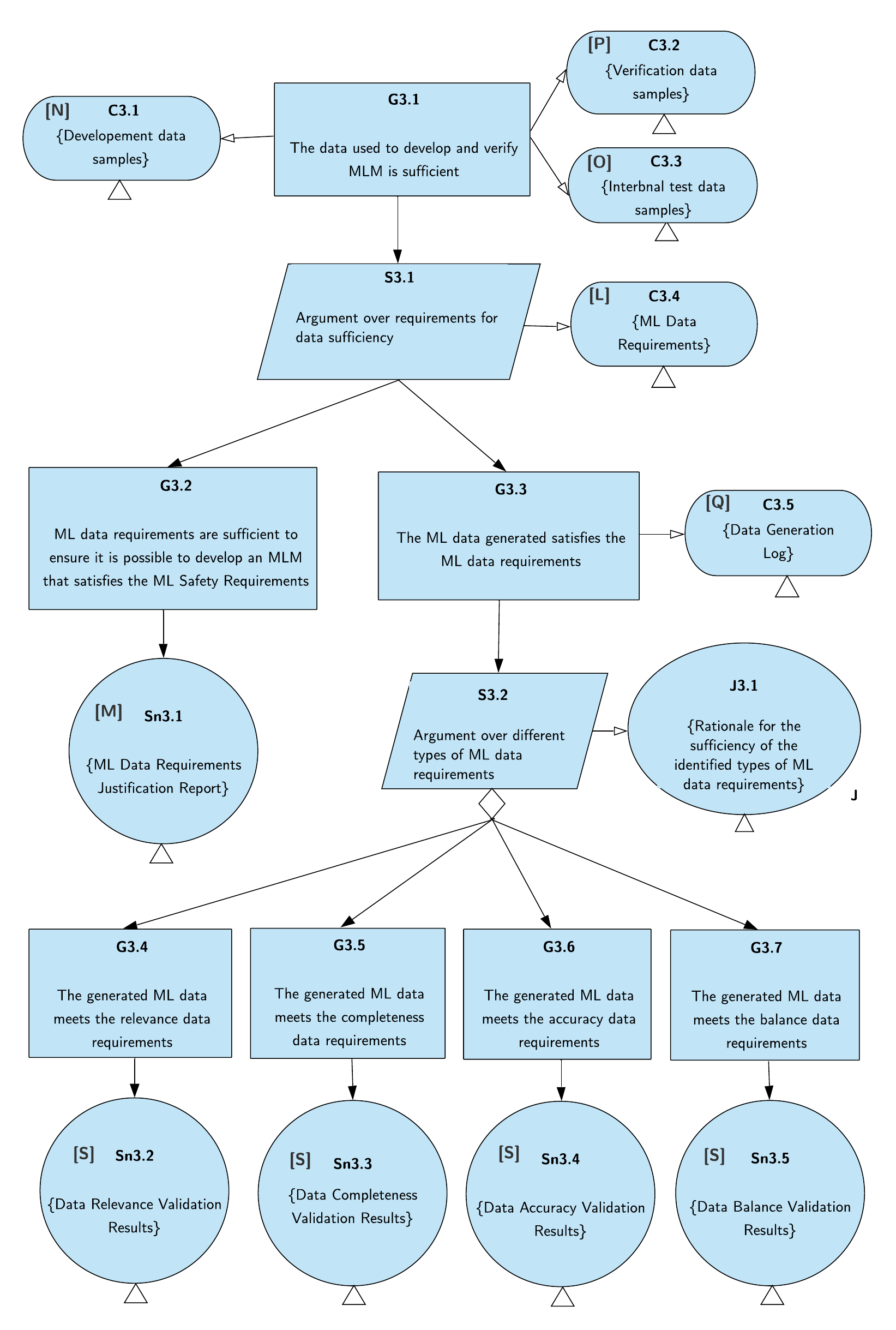}
    \caption{Assurance Argument Pattern for ML Data \ArtR}
    \label{fig:AArgMLD}
\end{figure}

The argument pattern relating to this stage of the AMLAS process is shown in Figure~\ref{fig:AArgMLD}. The key elements of the argument pattern are described below.

\subsection*{G3.1}
The top claim in this argument pattern is that the data used during the development and verification of the ML model is sufficient. This claim is made for all three sets of data used: development, test and verification (\ArtN, \ArtO, \ArtP). The argument sets out how the sufficiency of these data sets could be demonstrated. This provides confidence in the data used, and thus increases assurance of the model itself. 

\subsection*{S3.1}
The argument strategy is to argue over the defined ML data requirements which are provided as context to the argument (\ArtL). To support this strategy two sub-claims are provided in the argument, one demonstrating the sufficiency of the ML data requirements, and another to demonstrate that those defined data requirements are satisfied.

\subsection*{G3.2}
It is not possible to claim that the data alone can guarantee that the ML safety requirements will be satisfied, however the data used must be sufficient to enable the model that is developed to do so. This is shown by demonstrating that the requirements defined for the ML data are sufficient to ensure it is possible to create an ML model that satisfies the ML safety requirements. The ML Data Requirements Justification Report (\ArtM) created in Activity~\ref{act:defDR} is explicitly provided to provide evidence for this. 

\subsection*{G3.3}
It must be demonstrated that all of the data used throughout the lifecycle (development, test and verification) satisfies the defined ML data requirements. This is done in the context of the decisions made during data collection to ensure the data meets the requirements. These decisions are captured and explained in the data generation log (\ArtQ). 

To show that the data requirements are satisfied, the strategy adopted is to argue over each type of data requirement (relevance, completeness etc). The types of data requirements that have been considered should be justified. This is done explicitly in J3.1. 

For each type of data requirements, the ML data validation results (\ArtS) are used as evidence that each data set meets the requirements. 

\clearpage
\stage{Model Learning}

\subsection*{Objectives}
\begin{enumerate}
    \item Develop the machine learnt model using the development data obtained in the previous stage such that the allocated ML safety requirements are satisfied.  
    \item Use internal test data to assess the extent to which the machine learnt model is able to meet  the ML safety requirements when presented with data not used for development.                      
    \item Create an assurance argument, based on the evidence generated by meeting the first two objectives, which provides a clear justification that the ML model meets the  ML safety requirements. This should explicitly explain the tradeoffs, assumptions and uncertainties concerning both the ML model  and the process by which it is developed and validated.
\end{enumerate}

\subsection*{Inputs to the Stage}
\begin{itemize}
\item[\ArtH]: ML Safety Requirements
\item[\ArtN]: Development Data
\item[\ArtO]: Internal Test Data
\item[\ArtW]: ML Learning Argument Pattern 
\end{itemize}

\subsection*{Outputs of the Stage}
\begin{itemize}
\item[\ArtV]: ML Model
\item[\ArtX]: Internal Test Results
\item[\ArtY]: ML Learning Argument
\item[\ArtU]: Model Development Log
\end{itemize}

\subsection*{Description of the Stage}
\begin{figure}[htbp]
    \centering
    \includegraphics[width=0.7\linewidth]{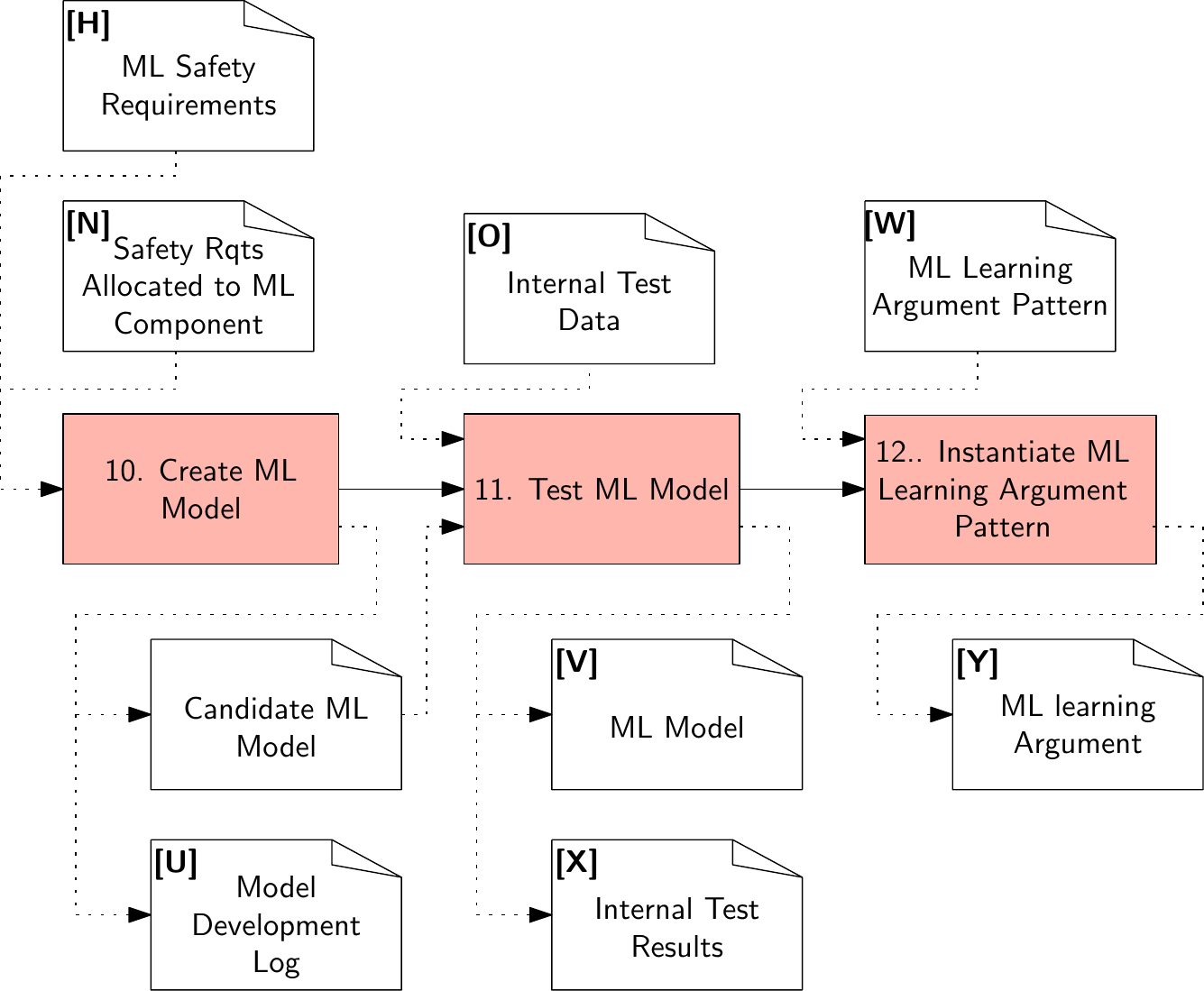}
    \caption{ AMLAS Model Learning  Assurance Process}
    \label{fig:MLProcess}
\end{figure}

As shown in Figure~\ref{fig:MLProcess},  this stage consists of three activities. The artefacts generated from this stage are used to instantiate the ML model assurance argument pattern as part of Activity~\ref{act:InstMLArgPattern}.

Additional guidance on this stage can be found at~\cite{AAIP-BoK-Learn}.

\Activity{Create ML Model \label{art:V} \ArtV}{act:CreateMLM}
An ML model meeting the ML Safety Requirements (\ArtH) shall be developed using the development data (\ArtN).  

The creation of an ML model starts with a decision as to the form of model that is most appropriate for the problem at hand and shall be most effective at satisfying the ML safety requirements. This decision may be based on expert knowledge and previous experience of best practice. The rationale shall be recorded in the model development log (\ArtU).

\begin{example}	Decision trees and random forests have been shown to provide excellent results for medical prognosis problems~\cite{martinez2017machine, dias2017risk, yue2018machine}. For low dimensional data they allow clinicians to understand the basis for decisions made by the system and, as such, may be more appropriate than neural networks for a range of problems.\end{example}

\begin{example}	Deep Neural Networks (DNNs) have the ability to extract features from image data. A DNN which receives images as a frame from a video feed has been shown to be capable of identifying objects in a scene and may therefore be suitable for use in an automotive perception pipeline.\end{example}

Typically numerous different candidate models of the selected type will be created from the development data by tuning the model hyperparameters in order to create models that may satisfy the ML safety requirements.

\begin{note}	A common problem that is encountered when creating a model is overfitting to training data. This happens when the model performs well using the development data but poorly when presented with data not seen before. This results from creating a model that focuses on maximising its performance for the data available, but whose performance does not generalise. Techniques such as cross-validation~\cite{anguita2012a}, leave-one-out~\cite{cheng2017a} and early stopping~\cite{prechelt1998a} can be used in handling the development data during the creation of the model in order to improve its generalisability and thus its ability to satisfy the ML safety requirements. \end{note}

\begin{example}	Let us consider a set of data samples which have been augmented with  samples from a realistic photo simulator. Those samples from the simulator have a higher proportion of hazardous scenarios since these are difficult/impossible to obtain from real-world experimentation. A model is then created to differentiate safe and hazardous situations. Both subsets of data are subject to noise, the first from  sensor noise the second from image artefacts in the simulator. Overfitting may occur when the model creates an overly complex boundary between two classes which aims to accommodate the noise present in those classes rather than the features which define the true class boundary.\end{example}

\begin{note}   In creating an acceptable model it is important to note that it is not only the performance of the model that matters. It is important to consider trade-offs between different properties such as trade-offs between cost of hardware and performance, performance and robustness or sensitivity and specificity. Several measures are available to assess some of these trade-offs. For example the Area Under ROC Curves enable the trade-offs between false-positive and false-negative classifications to be evaluated~\cite{fawcett2006a}. \end{note}

\subsection*{Artefact \ArtU:  Model Development Log}\label{art:U}
The process used in creating the model shall be documented in a model development log (\ArtU). The development log shall document and justify all key decisions made during the learning process (including the choice of development tool chain e.g. Tensorflow\footnote{https://www.tensorflow.org/} or pyTorch\footnote{https://pytorch.org/} machine learning platforms), and how those choices impact the performance or robustness of the model.

\begin{example}	The model development log may include details of model selection, changes to hyperparameters or changes to trade-off threshold definitions etc. For example “the threshold level for classification was set to x to ensure that the number of false positives identified in the development data was less than y”.
This information could aid developers when trying to develop a model which achieves acceptable false positive rates.
\end{example}

\begin{example}	In order to avoid overfitting and ensure that the model is able to perform in the presence of noise early stopping was employed. When the loss associated with model learning remained with $\epsilon$ for 10 iterations training was terminated.\end{example}

\begin{example} 	To ensure that the simplest model possible to meet the safety requirements was selected, regularisation is employed, adding a cost to the loss function to penalise the number of neurons in the model.\end{example}

\Activity{Test ML Model \label{act:V} \ArtV}{act:TestMLM}
Each candidate model created in Activity~\ref{act:CreateMLM} shall be evaluated using the internal test data (\ArtH) to check that it is able to satisfy the ML safety requirements. The internal test data shall not have been used during Activity~\ref{act:CreateMLM} in creating the candidate model\footnote{Allowing the development process to have a view of the internal test data is known as Data Leakage in Machine Learning~\cite{ashmore2019a}.}.

As shown in Figure~\ref{fig:MLProcess}, the model development stage is iterative and the model creation and model testing activities may be performed many times creating different models which will be evaluated in order to find the best one.  If it is not possible to create a model that meets the ML safety requirements with internal test data, the data management stage (Stage~\stageref{3}) and/or  the ML requirements stage (Stage~\stageref{2}) shall be revisited in order to create an acceptable model. Unlike traditional software testing, it is challenging to understand how an ML model can be changed to solve problems encountered during testing. The Model Development Log \ArtU~may provide insights to aid the developer to improve the model.

\begin{example}	In testing the model we find the accuracy is lower than expected indicating that the model fails to generalise beyond the development data. An analysis of the images that were incorrectly classified showed that images with bright sunlight have a higher failure rate than other images in the test set. This might dictate that we should return to the data management stage and collect additional images for this mode of failure. \end{example}

The results of the internal testing of the model shall be explicitly documented (\ArtX). 

A model shall be selected from the valid candidate models that have been created. The selected model (\ArtM) shall be the one which best meets the different, potentially conflicting, requirements that exist. This is a multi-objective optimisation problem where there could be multiple models on the pareto-front and it is important to select the best threshold to satisfy our requirements. 

\begin{example}	A model to be deployed in a perception pipeline classifies objects into one of ten classes. A set of ML safety requirements are defined in terms of the minimum accuracy for each class. The model development process returns 5 models each of which has accuracy greater than this minimum, however, each performs better with respect to one particular class. Under such conditions choosing the ‘best’ model requires the user to make a trade-off between class accuracies. Furthermore, this trade-off may change as we move from rural to urban contexts.\end{example}

\subsection*{Artefact \ArtX:  Internal Test Results}\label{art:X}
A document shall be created that records the results of executing the ML model (\ArtV) using the internal test data (\ArtO). The sufficiency of these results with respect to the defined ML safety requirements (\ArtH) shall also be documented.  

The measures used to report the internal testing results shall be appropriate for the defined ML safety requirements (\ArtH).

\begin{figure}[htbp]
    \centering
    \includegraphics[width=1\linewidth]{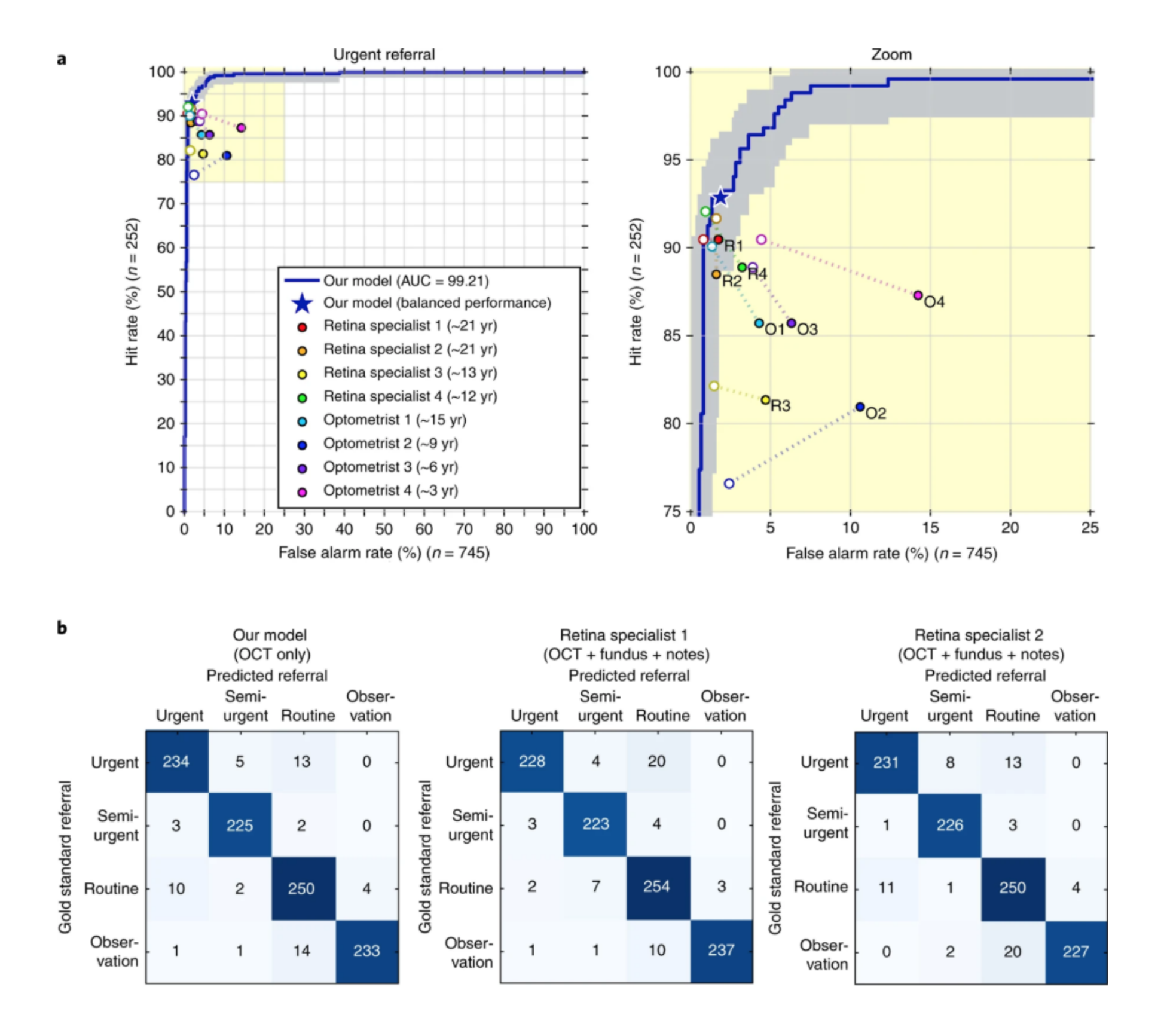}
    \caption{An example of model evaluation against different clinicians. Taken from~\cite{de2018a}}
    \label{fig:Results}
\end{figure}

\begin{example}	Figure~\ref{fig:Results} shows an example of performance evaluation for a system which is designed to diagnose patients for referral in retinal disease cases. The model classifies each case into one of four possible classes (urgent, semi-urgent, routine and observation). The costs associated with mis-classification are not uniform, e.g. misclassifying an urgent referral as observation carries a higher weight than misclassifying a semi-urgent referral as observation. For a single classifier model, (a) shows the true positive (hit rate) and false positive (false alarm) rates achieved for a set of test data as a threshold for identifying urgent referrals is varied.  Hence varying this threshold allows us to select a trade off which is acceptable for the system under test. This is presented in the form of a Receiver Operating Characteristic (ROC) curve \cite{fawcett2006a}. The right hand graph in figure~\ref{fig:Results} represents a detailed view of the highlighted area of the left hand ROC curve. The star marker on the ROC curve indicates the acceptable trade-off (``balanced performance'') for this system. We note that a perfect classifier would have zero false alarms and 100\% hit rate, this would be positioned in the top left corner of the graph.

In order to compare the performance of this model with the existing system, i.e. human experts, eight specialists were asked to classify the test cases and their performance was plotted on to the ROC curve. Where the specialist is to the right or below the ROC curve this indicates that the performance of the model is superior to the performance of the specialist.

While the ROC curve provides an aggregated assessment of model performance, a confusion matrix (b) \cite{geron2017hands} illustrates the model's misclassification at the class level such that we can observe the type of misclassification made. For example, in the left most matrix in figure~\ref{fig:Results} the ML model misclassified 'Urgent' as 'Routine' 13 times whilst in the central matrix, specialist 1 makes the same type of error 20 times. For this type of misclassification, the ML model can therefore be considered to out-perform specialist 1.

This example illustrates how multiple evaluation criteria may be required to be considered in order to assess the performance of an ML model for multi-objective problems such as the retinal diagnosis system considered. A trade-off is therefore required between different metrics in order to form a judgement on the acceptability of the model's performance in the given context of use. It is also noted that a single comparator may not be possible especially in socio-technical system such as this where human performance may vary considerably. 

\end{example}

\Activity{Instantiate ML Learning Argument Pattern \label{art:Y} \ArtY}{act:InstMLArgPattern}
This activity requires as input the ML learning argument pattern (\ArtW), as well as the artefacts from the previous activities of this stage (\ArtO, \ArtU, \ArtV~ and \ArtX). The activity uses these activities and outputs from the previous stages to create an instantiated ML learning argument (\ArtY).

\subsection*{Artefact \ArtW: Model Learning argument pattern}\label{art:W}
\begin{figure}[htbp]
    \centering
    \includegraphics[width=0.9\linewidth]{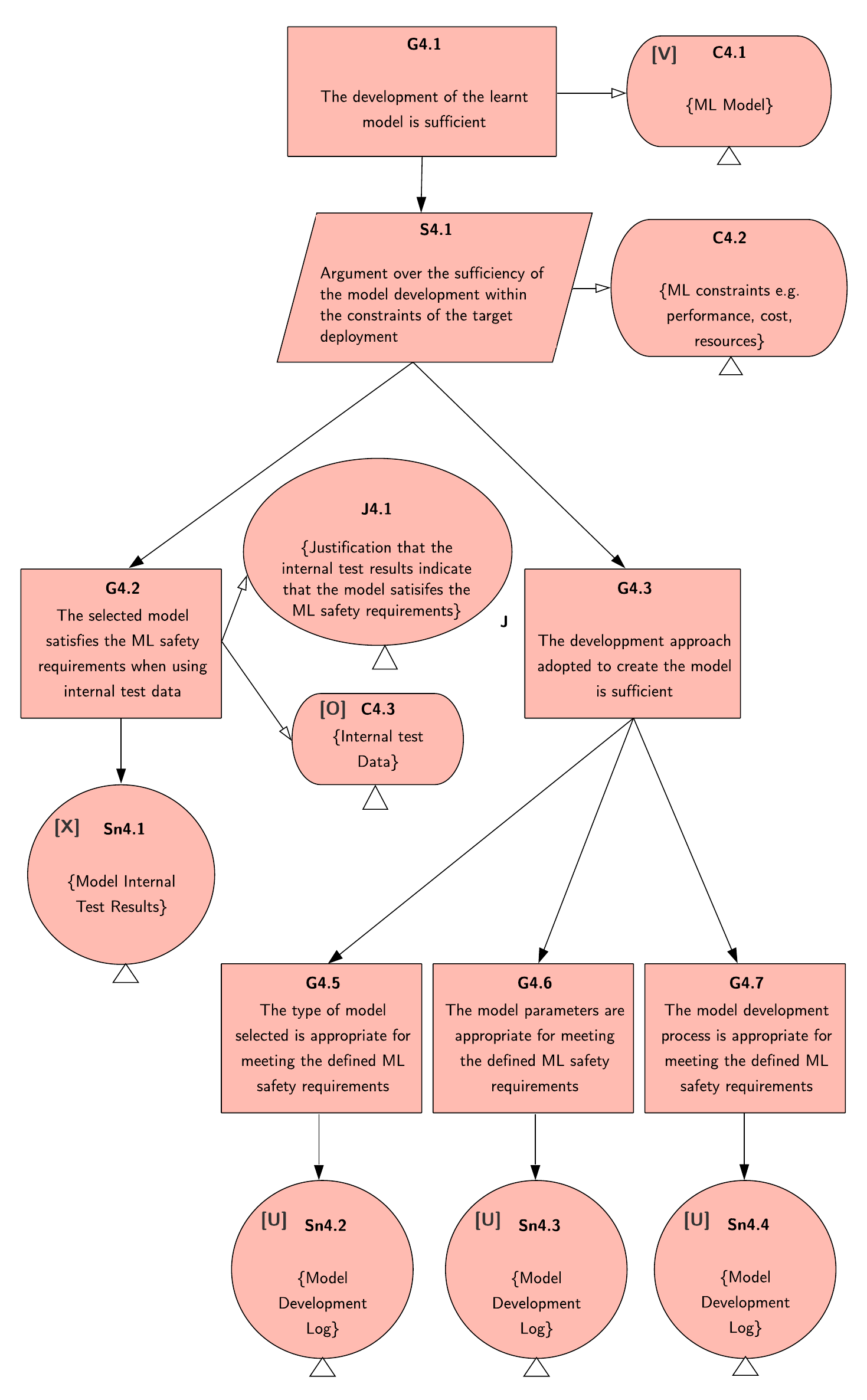}
    \caption{Assurance Argument Pattern for Model Learning}
    \label{fig:ArgPatML}
\end{figure}

The argument pattern relating to this stage of the AMLAS process is shown in Figure~\ref{fig:ArgPatML}. The key elements of the argument pattern are described below. 

\subsection*{G4.1}
The top claim in this argument pattern is that the development of the learnt model (\ArtV) is sufficient. The sufficiency of the model learning process is argued through considering the appropriateness of the model development activities undertaken.

\subsection*{S4.1}
The argument strategy is to argue over the internal testing of the model performed during development as well as the development approach adopted. The appropriateness of the development activities is considered within the context of creating a model that both satisfies the ML safety requirements as well as meeting the additional constraints that are imposed on the model, such as performance and cost.

\subsection*{G4.2}
It must be demonstrated that the ML model that is selected satisfies the ML safety requirements. This is shown by using the internal test data (\ArtO) generated from Activity~\ref{act:GenMLD}. The model must be shown to satisfy the ML safety requirements when this test data is applied. The internal testing claim is supported through evidence from the internal test results (\ArtX).

A justification must be provided that the results obtained from the internal testing are sufficient to indicate that the ML safety requirements are satisfied. This justification is provided in J4.1. 

\subsection*{G4.3}
This claim considers the approach that has been adopted in developing the model. This claim is supported by claims regarding the type of model selected, the model parameters that are used and the process that is applied.

\subsection*{G4.5}
It must be demonstrated that the type of model that is created in Activity~\ref{act:CreateMLM} is appropriate for the given set of ML safety requirements and the other model constraints. The evidence for the type of model selected is captured in the model development log (\ArtU), which is used as evidence to support this claim.

\subsection*{G4.6}
It must be demonstrated that the parameters of the selected model have been appropriately tuned in Activity~\ref{act:CreateMLM}. The parameters must be shown to be appropriate for the given set of ML safety requirements. The rationale for how the model parameters are determined should be captured in the model development log (\ArtU), which is used as evidence to support this claim.

\subsection*{G4.7}
It must be demonstrated that the process is appropriate. As discussed in Activity~\ref{act:CreateMLM}, this will be a highly iterative process involving a number of decisions on each iteration, and the development of multiple models. The process will also involve decisions regarding the model architecture. The rationale for the process decisions should be included in the model development log (\ArtU) along with a justification for the appropriateness of the development tool chain used.

\clearpage
\stage{Model Verification}

\subsection*{Objectives}
\begin{enumerate}
\item Demonstrate that the model will meet the ML safety requirements when exposed to inputs not present during the development of the model.
\item Create an assurance argument, based on the evidence generated by the first objective. The argument should clearly demonstrate the relationship between the verification evidence and the ML safety requirements. It should explicitly explain the tradeoffs, assumptions and uncertainties concerning the verification results  and the process by which they were generated.
\end{enumerate}

\subsection*{Inputs to the Stage}
\begin{itemize}
\item[\ArtH]: ML safety requirements
\item[\ArtP]: Verification data
\item[\ArtV]: ML Model
\item[\ArtAB]: ML verification argument pattern
\end{itemize}

\subsection*{Outputs of the Stage}
\begin{itemize}
\item[\ArtZ]: ML verification results
\item[\ArtAA]: Verification log
\item[\ArtAC]: ML Verification argument
\end{itemize}

\subsection*{Description of the Stage}
\begin{figure}[h]
    \centering
    \includegraphics[width=0.7\linewidth]{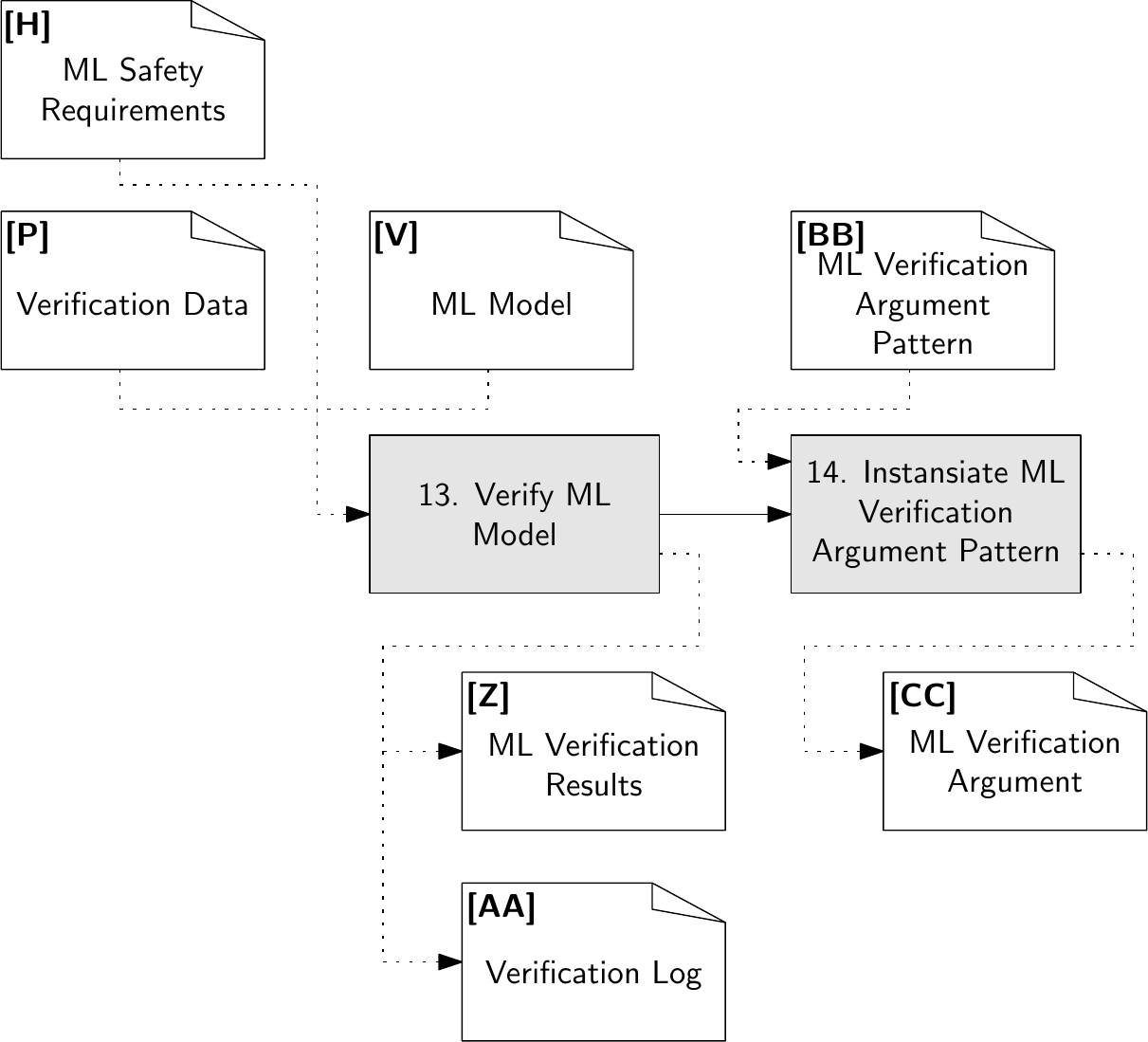}
    \caption{ AMLAS Model Verification Assurance Process\label{fig:VerProcess}}
\end{figure}

As shown in Figure~\ref{fig:VerProcess}, this stage consists of two activities that are performed to provide assurance in the ML Model verification process. The primary artefacts generated from this stage are ML model verification results which are used to instantiate the ML verification argument pattern as part of Activity~\ref{act:InstMLVer}.

Additional guidance on this stage can be found at~\cite{AAIP-BoK-Verif}.

\Activity{Verify ML Model}{act:Verify}
This activity requires as input the ML safety requirements (\ArtH), the verification data (\ArtP) and the machine learnt model (\ArtV). Model verification may consist of two sub-activities: test-based verification and formal verification. For every ML safety requirement at least one verification activity shall be undertaken. The results of verification for each requirement shall be recorded in the ML verification results (\ArtZ).

All verification activities shall be sufficiently independent from the development activities. A log (\ArtAA) shall be created that documents the measures taken to verify the ML model, including those measures taken to ensure that data used in verification was not exposed to the development team. 

\begin{note}	Independence should be proportionate to the criticality of the ML model. In addition to different degrees of independence between the development and verification data sets e.g. data sets compiled from one or more hospitals, further independence may be necessary between the development and verification engineers at the team or organisational levels.
\end{note}

One of the aims of model verification is to show that the performance of the model with respect to ML safety requirements encoded as metrics such as precision and recall are maintained when the model is subjected to inputs not present in the development data. A model which continues to perform when presented with data not included in the development set is known in the ML community as generalisable. Failures to generalise can be due to a lack of feature coverage in the development data or a lack of robustness to those perturbations which may be considered to be noise i.e. small changes to a data sample which meets the performance specification in the absence of such noise.

\begin{em}Test-based verification\end{em} utilises the verification data to demonstrate that the model generalises to cases not present in the model learning stage. This shall involve an independent examination of the properties considered during the model learning stage. Specifically, those safety requirements associated with ensuring the robustness of models are evaluated on the independent verification data set i.e. that the performance is maintained in the presence of adverse conditions or signal perturbations. The test team should examine those cases which lie on boundaries or which are known to be problematic within the context to which the model is to be deployed. 

\pagebreak
If the nature of the model results in a verification test which is unable to determine if the model satisfies the safety requirement, it may be necessary to augment the verification data set to demonstrate definitively if the requirement is met. This may for example arise due to non-linearities in a model.

\begin{example}
A neural network which predicts the stopping distance of a vehicle given environmental conditions is tested across a range of linearly spaced values (temperature, humidity, precipitation). The results show that the model operates safely at all values except 3 distinct points in the range. Further verification data may need to be gathered around these points to determine the exact conditions leading to  safety violations.
\end{example}

\begin{note} 	It is important to ensure that the verification data is not made available to the development team since if they are to have oversight of the verification data this is  likely to lead to techniques at development time which circumvent specific samples in the verification set rather than considering the problem of generalisation more widely.
\end{note}

\begin{note}	The ML verification results should evaluate test completeness with respect to the dimensions of variability outlined in the ML safety requirements. This is directly related to the desire for data completeness outlined in Stage~\stageref{3} of the process.
\end{note}

\begin{example} 	Since we know that material on a camera lens can lead to blurring in regions of an image, we may make use of ‘contextual mutators’~\cite{pezzementi2018a} to assess the  robustness of a neural network with respect to levels of blur. In this way the level of blur which can be accommodated can be assessed and related to contextually meaningful measures.
\end{example}

\begin{note}	For some safety-related properties, such as interpretability, it may be necessary to include a human in the loop evaluation mechanism. This may involve placing the component into the application and generating explanations for experts to evaluate~\cite{doshi-velez2017a} \end{note}

\begin{example}	The DeepMind retinal diagnosis system~\cite{yim2020a} generates a segmentation map as part of the diagnosis pipeline. This image may be shown to clinical staff to ensure that the end user is able to understand the rationale for the diagnosis. As part of a verification stage these maps may be presented to users without the associated diagnosis to ensure that the images are sufficiently interpretable.\end{example}

\begin{em}Formal verification\end{em} uses mathematical techniques to prove that the learnt model satisfies formally-specified properties derived from the ML safety requirements. When formal verification is applied, counter-examples are typically created which demonstrate the properties that are violated. In some cases, these may be used to inform further iterations of requirements specification, data management or model learning.

The formally-specified properties shall be a sufficient representation of the ML safety requirements in the context of the defined operating environment. An explicit justification shall be documented for the sufficiency of the translation to formal properties. 

\begin{note}	The number of formal methods available for machine learnt models is increasing rapidly. However, it is not always possible to map formal results to a context which is easily understood by those wishing to deploy the system. Techniques which guarantee point-wise robustness for individual samples~\cite{mohapatra2019a}, for example, are mathematically provable. For high dimensional input spaces, such as images however, knowing that a network is robust for a single sample if the pixel values change by less than 0.06 does little to inform the user as to the robustness of the image in rain.\end{note}

\begin{example}	Formal verification of neural networks are able to demonstrate that a perturbation of up to threshold value, $\epsilon$, on all the inputs leads to a classification that is the same as the original sample class~\cite{katz2017b,katz2019a}. Such a result is only meaningful when the value of $\epsilon$ is translated into contextually meaningful values such as steering angles. It is of little value when it simply provides a range of variation in pixel values for a single input sample (point wise verification).\end{example}

\begin{example}	An airborne collision avoidance system, AcasXu~\cite{katz2017a}, provides steering advisories for unmanned aerial vehicles. A neural network implementation of the system was verified using satisfiability modulo theories (SMT) solvers to ensure that alerting regions are uniform and do not contain inconsistent alerts.\end{example}

The formal models that are used for verification will require assumptions and abstractions to be made, both with respect to the ML model itself, and with respect to the operating environment. The validity of the formal model shall therefore be demonstrated~\cite{habli2009a}.

\subsection*{Artefact \ArtZ : ML verification evidence}\label{art:Z}
Having undertaken verification activities, ML verification evidence should be collated and reported in terms which are meaningful to the safety engineer with respect to the ML safety requirements and the operating environment. 
The verification evidence shall be comprehensive and shall clearly demonstrate coverage with respect to the dimensions of variability, and combinations thereof, relevant to the ML safety requirements.

\subsection*{Examples of verification evidence from testing:}

\begin{example}	DeepXplore~\cite{pei2017a} and Deep Test~\cite{tian2018a} provide evidence which can be thought of as  similar to software coverage tests for Deep Neural Networks. This technique considers the firing of neurons in a network when exposed to the data samples in the training set. While this is not strictly analogous to traditional coverage measures it allows for the  identification of those input samples which are ‘corner-cases’ with respect to the training data.
\end{example}

\begin{example} 	Deep Road~\cite{zhang2018a} utilises Generative Adversarial Network (GAN) based techniques to synthesize realistic and diverse driving scenarios in order to find inconsistencies in autonomous driving systems. The evidence should enumerate the scenarios examined and the results of the model when presented with these samples as well as the ground truth labels.\end{example}

\subsection*{Examples of verification evidence from formal verification:}

\begin{example}	Local Robustness Property~\cite{huang2017a}: Given a deep neural network and an input region defined by a data sample, local robustness defines a region around the input for which the output of the function defined by the neural network remains unaltered. Evidence is then the samples verified and the dimension of the region for which robustness is proved.\end{example}

\begin{example} 	DeepSafe~\cite{gopinath2017a} identifies ‘safe regions’ in the input space for classification problems. The input space is first partitioned into regions which are likely to have the same label. These regions are then checked for targeted robustness i.e. no point in the space can be mapped to a specific incorrect label. More concretely, when considering an unmanned aircraft control system DeepSafe was able to identify regions for the inputs to the system for which a ‘weak left’ turn advisory would always be returned.\end{example}

\subsection*{Artefact \ArtAA: Verification Log}\label{art:AA}

This log should explicitly document the verification strategy. For testing this should include the range of tests undertaken and the rationale for performing each test with bounds and test parameters where appropriate. In addition the approaches taken to manage verification data in such a way as to ensure that data leakage did not occur should be documented. For formal verification the techniques employed should be listed and the rationale for using such approaches to verify properties of the model with respect to real-world features included.

\begin{note}	The verification of properties using formal techniques typically require requirements to be encoded using a restricted language set e.g. First order logics, or domain specific languages. Converting natural language requirements into these representations may require significant domain expertise and shall be explicitly documented in the verification log.\end{note}

\Activity{Instantiate ML Verification Argument Pattern \label{art:CC} \ArtAC}{act:InstMLVer}

This activity requires as input the ML verification argument pattern (\ArtAB), as well as the artefacts from the previous activities of this stage (\ArtAA, \ArtZ and \ArtP). The activity uses these activities and outputs from the previous stages to create an instantiated ML verification argument (\ArtAC).

\subsection*{Artefact \ArtAB: ML Verification Argument Pattern}\label{art:BB}

\begin{figure}[htbp]
    \centering
    \includegraphics[width=1\linewidth]{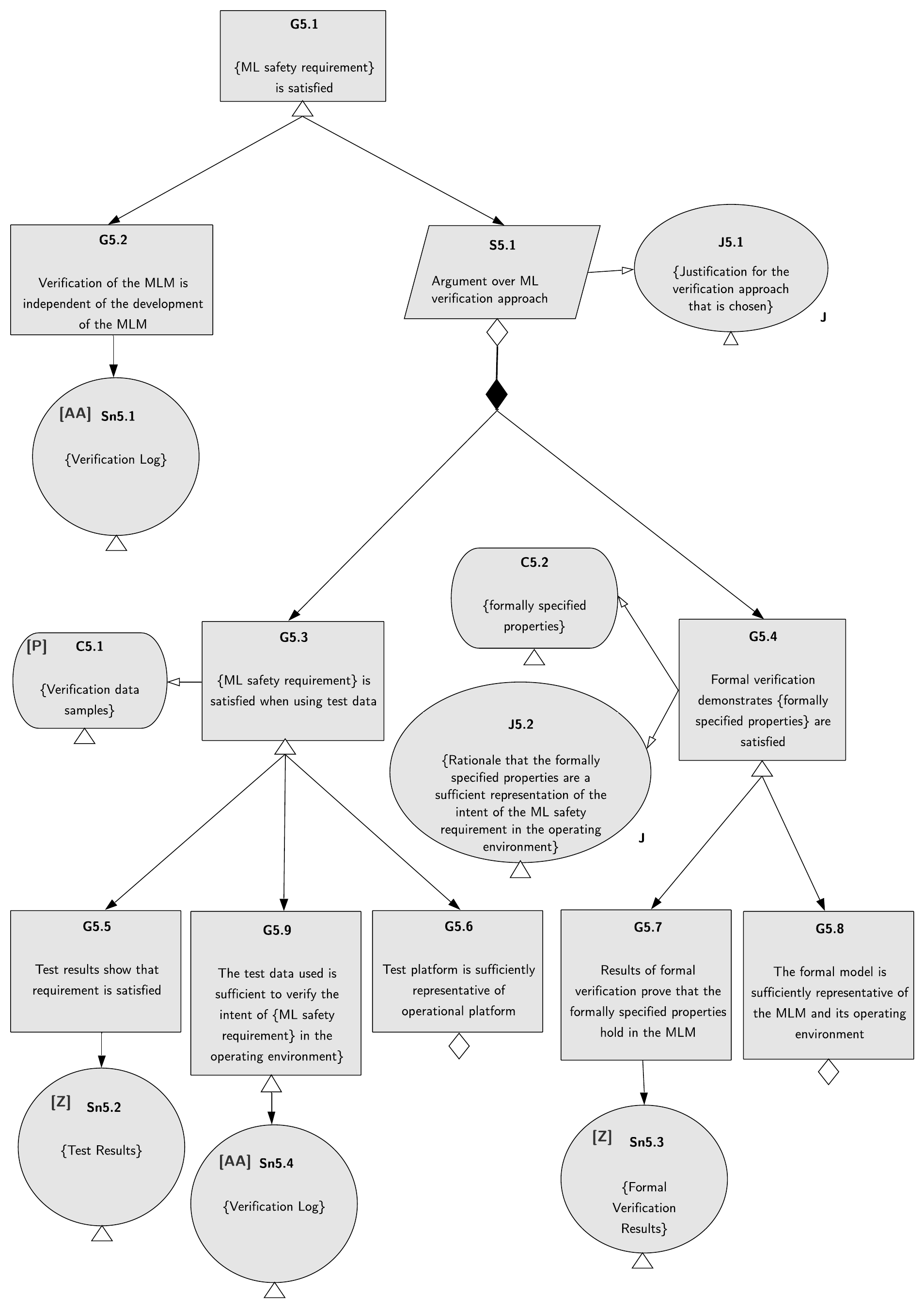}
    \caption{Assurance Argument Pattern for ML Verification  }
    \label{fig:ArgPatMLVer}
\end{figure}

The argument pattern relating to this stage of the AMLAS process is shown in Figure~\ref{fig:ArgPatMLVer}. The key elements of the argument pattern are described below.

\subsection*{G5.1}
The top claim in the verification argument pattern corresponds to the bottom claim in the safety requirements argument pattern (\ArtI); it is at this point that each ML safety requirement that has been established must be shown to be met. The satisfaction of the requirement is shown through the verification activities that are performed, as discussed in Activity~\ref{act:Verify}. This claim is supported by strategy S5.1 that reasons about the verification activities undertaken and a claim G5.2, that provides evidence from the Verification log (\ArtAA) that the verification activities have been performed independently from the development of the ML model.

\subsection*{S5.1}
In order to demonstrate that the ML safety requirement is sufficiently satisfied, the pattern provides a choice over how the claim can be supported. The evidence may come, as discussed in Activity~\ref{act:Verify}, from any combination of testing and formal verification. The choice in the argument should be interpreted as “at-least-1”, allowing for multiple legs of argumentation. The combination of verification approaches used should be justified in J5.1. The ``requires development'' adornment to strategy S5.1 indicates that other verification approaches may optionally also be adopted where this is felt to be required. An argument and evidence regarding any such approaches must be included in the assurance argument.

\subsection*{G5.3}
When the verification strategy includes test-based verification, it must be demonstrated that the ML model satisfies the ML safety requirement when the verification data is applied. The testing claim is supported through evidence from the test results (\ArtZ). For any ML safety requirement, the test data used will be a subset of the verification data samples (\ArtP) generated from Activity~\ref{act:GenMLD}. The test data must demonstrate that the ML safety requirement is satisfied across a sufficient range of inputs representing the operating environment, that are not included in the data used in the model learning stage. The sufficiency of the test data is justified in the verification log (\ArtAA).  It is also necessary to consider the way in which the test results were obtained. This is particularly important where testing is not performed on the target system. This is considered in G5.6 where evidence must be provided to demonstrate that the test platform and test environment used to carry out the verification testing is sufficiently representative of the operational platform of the system to which the ML component will be deployed. G5.6 is not developed further as part of this guidance. 

\subsection*{G5.4}
When the verification strategy includes formal verification, a claim is made that the ML model satisfies formally specified properties. The formally specified properties should be a sufficient formal representation of the intent of the ML safety requirement that is being verified. A justification should be provided in J5.2 to explain the sufficiency of the translation from the ML safety requirement to the formally specified properties. The formal verification claim is supported through evidence from the formal verification results (\ArtZ). For those results to be valid, it must be demonstrated that the formal model created to perform the verification is sufficiently representative of the behaviour of the learnt model, and that all assumptions made as part of the verification about the system and operating environment are valid. This argument is made under G5.8, which is not developed further as part of this guidance.

\clearpage
\stage{Model Deployment}

\subsection*{Objectives}
\begin{enumerate}
    \item Integrate the machine learnt component into the target system in such a manner that the system satisfies the allocated system safety requirements. The component should be integrated in the pipeline linking its inputs and outputs to other system components.
    \item Demonstrate that the allocated system safety requirements are still satisfied during operation of the target system and environment.
    \item Create an assurance argument to demonstrate that the ML model will continue to meet  the  ML safety requirements once integrated into the target system. 
\end{enumerate}

\subsection*{Inputs to the Stage}
\begin{itemize}
    \item[\ArtA]: System Safety Requirements
    \item[\ArtB]: Environment Description
    \item[\ArtC]: System Description
    \item[\ArtV]: ML Model
    \item[\ArtAG]: ML Deployment Argument Pattern
    \item[\ArtAE]: Operational scenarios
\end{itemize}

\subsection*{Outputs of the Stage}
\begin{itemize}
    \item[\ArtAD]: Erroneous Behaviour Log 
    \item[\ArtAF]: Integration Testing Results
    \item[\ArtAH]: ML Deployment Argument
\end{itemize}

\subsection*{Description of the Stage}
\begin{figure}[h]
    \centering
    \includegraphics[width=1\linewidth]{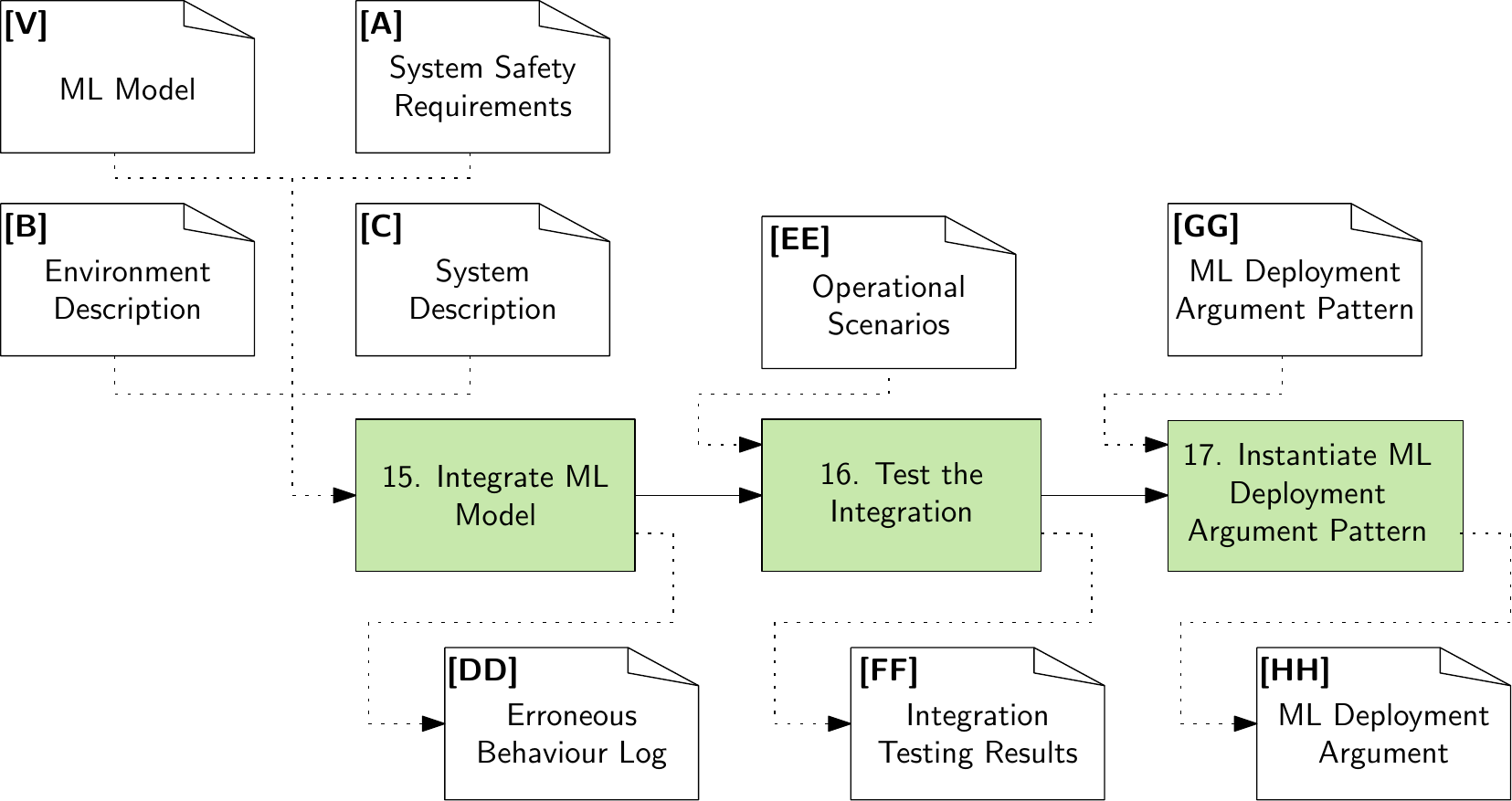}
    \caption{ AMLAS Model Deployment Assurance Process  }
    \label{fig:DeploymentProcess}
\end{figure}

As shown in Figure~\ref{fig:DeploymentProcess},  this stage consists of three activities that provide a basis for ML component deployment assurance. This process shall be followed not only for initial deployment of the component but also for any subsequent deployment required to update the component within the system. The artefacts generated from this stage are used to instantiate the ML model deployment assurance argument pattern as part of Activity~\ref{act:InstDeployPattern}.

Additional guidance on this stage can be found at~\cite{AAIP-BoK}.

\Activity{Integrate ML Model}{act:Integrate}
The ML Model needs to be deployed onto the intended hardware platform and integrated into the broader system of which it is a part. 
Deploying the component may be a multi-stage process in which the component is first deployed to computational hardware which is then integrated at a subsystem level before being integrating with the final hardware platform.
The deployment process  will include, connecting the component's  inputs to sensing devices (or equivalent components) and providing its output to the wider system. This activity takes as inputs the system safety requirements (\ArtA), the environment description (\ArtB), the system description (\ArtC) and the ML model (\ArtV) defined in the previous stages and integrates the model into the overall system .  

The development of the ML model is undertaken in the context of assumptions that are made about the system to which the ML model will be integrated (\ArtC) and the operating environment of that system (\ArtB). This will include key assumptions that, if they do not hold during operation of the system, may result in the ML model not behaving in the manner expected as a result of development and verification activities.

\begin{example}	When an ML component used for object classification is developed an assumption is made that the component will only be used in good lighting conditions. This may be based on the capabilities of the sensors, historic use cases, and the data from which the component is trained and verified. It is crucial to recognise and record that this is a key assumption upon which the assurance of the ML component is based. If a system containing the component is subsequently used at low light levels then the classification generated by the ML component may not meet it's safety requirements.
\end{example}

\begin{note} 	When considering violations of assumptions, this should be linked to the system safety analysis process to identify the impact on system hazards and associated risks.\end{note}

Measures shall be put in place to monitor and check validity throughout the operation of the system of the key system and environmental assumptions. Mechanisms shall be put in place to mitigate the risk posed if any of the assumptions are violated. Further guidance on the deployment of components to autonomous systems may be found in~\cite{SASWG2020, ashmore2019a}.

\begin{example} 	There is an assumption that the ML component for pedestrian detection deployed in a self driving car will be used only in daylight conditions. The system monitors the light levels. If the level of light drops below a level defined in the operating environment description then the car shall hand back control to a human driver.\end{example}

\begin{note} 	Monitoring the systems is an ongoing runtime activity. Inputs can be monitored with appropriate statistical techniques to make sure that they are close to the training data distributions. In some cases the model itself can give to the system a value representing its confidence in the output~\cite{waa2018a}.  The human can be added to the feedback to help audit model inputs and environment.\end{note}

There will always be some level of uncertainty associated with the outputs produced by any ML model that is created. This uncertainty can lead to erroneous outputs from the model. The system shall monitor the outputs of the ML model during operation, as well as the internal states of the model, in order to identify when erroneous behaviour occurs. These erroneous outputs, and model states, shall be documented in the erroneous behaviour log (\ArtAD)

As well as considering how the system can tolerate erroneous outputs from the ML model, integration shall consider erroneous inputs to the model. These may arise from noise and uncertainties in other system components; as a result of the complexity of the operating environment; or due to adversarial behaviours. These erroneous inputs shall be documented in the erroneous behaviour log (\ArtAD).

\begin{example} 	Due to occlusion of a pedestrian in an image due to other objects in the environment a pedestrian may briefly not be detected. We know that in the real world a human does not disappear, so the system can  use this knowledge to ignore non-detections of previously identified pedestrians that last for a small number of frames.\end{example}

When integrating the model into the system the suitability of the target hardware platform shall be considered [54]. During the development of the model, assumptions are made about the target hardware and the validity of those assumptions shall be checked during integration. If the target hardware is unsuitable for the ML model then a new model may need to be developed.

\begin{example}	 It may be possible to create a  complex deep neural network that provides excellent performance. However such a model might require large computational power to execute.  If the hardware in which the model will be deployed does not have sufficient computational power then a different model may need to be created in order to reduce the required computational power. \end{example} 

\begin{note} It is important to evaluate latency associated with accessing input data. Relying on sensing data from other systems, via external networks, may unacceptably slow down the output of the ML component.\end{note}

The system in which the ML model is deployed shall be designed such that the system maintains an acceptable level of safety even in the face of the predicted erroneous outputs that the model may provide. 

\begin{example}	The ML model for pedestrian detection deployed in a self driving car has a performance requirement of 80\% accuracy. Due to uncertainty in the model this performance cannot be achieved for every frame. The model uses as inputs a series of multiple images derived from consecutive image frames obtained from a camera.  The presence of a pedestrian is determined by considering the result in the majority of the frames in the series.  In this way the system compensates for the possible error of the model for any single image used. \end{example}

\subsection*{Artefact \ArtAD: Erroneous Behaviour Log}\label{art:DD}

 The  nature and characteristics of the erroneous outputs shall be predicted and documented in the erroneous behaviour log (\ArtAD) such that an appropriate system response can be determined. These predictions shall be informed by the findings of internal testing and of verification activities. This understanding can be enhanced through integration testing performed at Activity~\ref{act:TestIntegration}.

\Activity{Test the Integration \label{art:FF} \ArtAF}{act:TestIntegration}
Once the ML model has been integrated into the wider system, the integration needs to be tested to check that the system safety requirements (\ArtA) are satisfied.  This activity requires a defined set of operational scenarios (\ArtAE) against which the behaviour of the system, as implemented in ML, can be tested.

\subsection*{Artefact \ArtAE: Operational Scenarios}\label{art:EE}
An operational scenario is defined as “Description of an imagined sequence of events that includes the interaction of the product or service with its environment and users, as well as interaction among its product or service components”~\cite{i2011a}. The set of operational scenarios shall therefore represent real scenarios that may be encountered when the system is in operation. This set shall comprise a number of defined  scenarios, meaningful with respect to the safety requirements of the system, that may occur during the system life. 

\begin{example}  For a system designed for automated retinal disease diagnosis an operational scenario can be represented using relevant clinical pathways.\end{example}

\begin{example}  For a self driving car, different operational scenarios would cover changing lanes on a motorway, navigating a roundabout, stopping at a red light etc.\end{example}

The system shall be tested against the defined operational scenarios (\ArtAE), and the results from the tests assessed against the safety requirements. The results shall be captured explicitly as the integration testing results (\ArtAF). 

Integration testing may take many forms, including  simulation and hardware in the loop testing~\cite{bjelevac2015a}.  A shadow deployment can also be used for integration testing to evaluate the actual system in the real operating environment while another stable system is in use.

When using simulation, sufficient confidence shall be demonstrated that the simulator represents the actual operating environment.

\begin{example} Implementing SLAM (simultaneous localization and mapping) using a simulated environment such as gazebo\footnote{http://gazebosim.org/} can provide very different results from considering a physical robot such as a turtlebot\footnote{https://emanual.robotis.com/docs/en/platform/turtlebot3/overview/} navigating in a real-world setting. In particular, simulators often do not take good account of effects of sensor noise such as reflections in the case of LIDAR.
\end{example}

\begin{note}	Not all properties of the system can be tested in simulation. Physical properties such as different frictions depending on wearing or sensor noises are very difficult to simulate and will require hardware in the loop (HIL) or real system testing. HIL has the advantage of enabling testing in the early stages of a project and can be used to take account of sensor noise or other physical properties in different programmed virtual environments.\end{note}

The target system containing the integrated ML component shall be tested  in a controlled setting to allow for safe evaluation of the system. This controlled setting may include additional controls,  monitoring, or the use of simulation of real-world scenarios.
In this way the  behaviour of the component may be safely evaluated, by stakeholders, in context.

\begin{example} An automated retinal disease diagnosis system is used as a trial in an hospital for six months providing guidance to clinicians who make the diagnosis decisions. This approach allows the clinicians to evaluate the performance of the system whilst maintaining the ability to override any diagnosis results from the system that are felt to be unsafe. \end{example}

\begin{example} Prior to operation on public highways, an automated lane following system may be tested and tuned once integrated into a vehicle making use of experienced test drivers and on designated test tracks. This reduces the risk during evaluation of the system in high speed manoeuvres.\end{example}

Wherever possible the integration to the system shall be tested using the actual target system or using a hardware in the loop approach, as this provides results that most closely reflect what will be observed in operation. However  in many cases this may be impractical. In which case simulation and hardware in the loop may be used together. 

The integration testing results shall be reported in the integration test results (W) artefact providing evidence that the system safety requirements (A) are met. 

\begin{example} Worst case execution time of a system used for a vehicle control loop which includes a machine learning component shall remain in the limit for each of the different scenarios in order to assure the safety of the vehicle.\end{example}

\Activity{Instantiate ML Deployment Argument Pattern\label{art:HH} \ArtAH}{act:InstDeployPattern} 

This activity requires as input the ML verification argument pattern (\ArtAB), as well as the artefacts from the previous activities of this stage (\ArtAA, \ArtZ and \ArtP). The activity uses these activities and outputs from the previous stages to create an instantiated ML verification argument (\ArtAC).

\subsection*{Artefact \ArtAG: ML Deployment Argument Pattern}\label{art:GG}

\begin{figure}[htbp]
    \centering
    \includegraphics[width=1\linewidth]{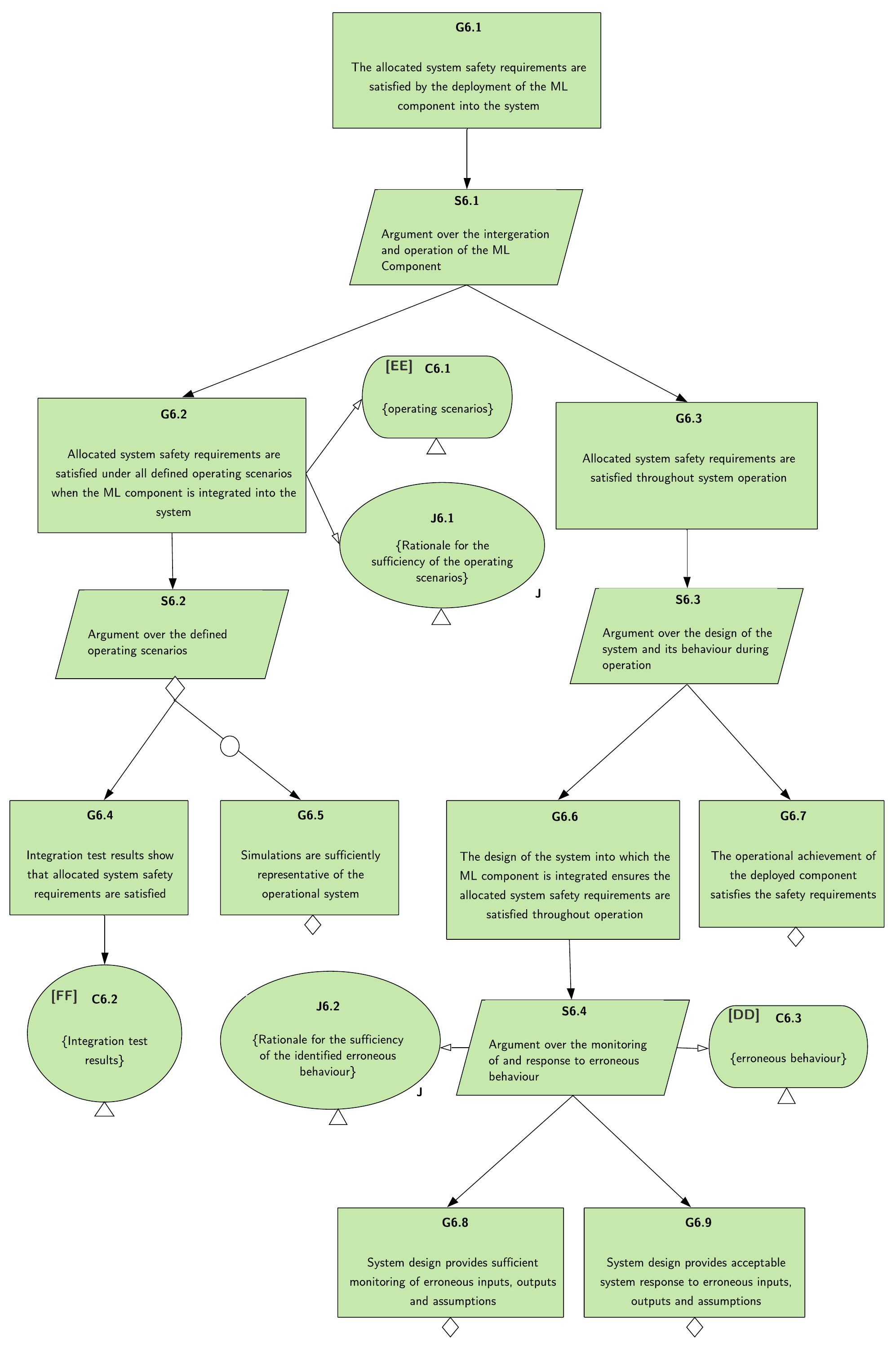}
    \caption{ Assurance Argument Pattern for ML Model Deployment }
    \label{fig:ArgPatDeployment}
\end{figure}

The argument pattern relating to this stage of the AMLAS process is shown in Figure~\ref{fig:ArgPatDeployment}. The key elements of the argument pattern are described below.

\subsection*{G6.1}
It must be demonstrated that the safety requirements allocated to the ML component are still met when the ML component is deployed to the system in which it operates. This is shown by providing two sub-claims. Firstly, the ML component integration claim demonstrates that the safety requirements (that were satisfied by the ML model) are also met when the ML component is integrated into the rest of the system. Secondly, the ML component operation claim is provided to show that the safety requirements will continue to be met throughout the operation of the system.

\subsection*{G6.2}
It must be demonstrated that the safety requirements allocated to the ML component are satisfied when the component is integrated to the system. To demonstrate this, the ML component must be executed as part of the system following integration. It must be checked that the safety requirements are satisfied when the defined set of operating scenarios are executed. The operating scenarios used in the integration testing (\ArtAF) are provided as context for the claim.  The sufficiency of the operating scenarios that are used must be justified in J6.1. This justification explains how the scenarios were identified such that they represent real scenarios of interest that may be encountered when the system is in operation. 

\subsection*{S6.2}
The strategy to support the integration claim is to firstly use the integration test results (\ArtAF) to demonstrate the safety requirements are met for the defined operating scenarios. Integration testing is often performed for autonomous systems using a simulator. Where this is the case it is also necessary to demonstrate that the simulations that are used are a sufficient representation of the operational system to which the ML component is deployed. Evidence for this will be provided to support claim G6.5.

\subsection*{G6.3}
It must also be demonstrated that the safety requirements allocated to the ML component continue to be satisfied during the operation of the system. To demonstrate this, claim G6.6 shows that the system is designed such that it supports the safe operation of the ML component, and G6.7 demonstrates that the observed behaviour during operation continues to satisfy the safety requirements. In a complete safety case for an ML component argument and evidence to support this claim would be required, further guidance on this is provided in~\cite{AAIP-BoK-Main}.

\subsection*{G6.6}
It must be demonstrated that the design of the system into which the ML component is integrated is robust by taking account of the identified potential erroneous behaviour (\ArtAD). It must be shown that predicted erroneous behaviour will not result in violation of the safety requirements. In particular the argument must focus on erroneous inputs to the ML component from the rest of the system and erroneous outputs from the ML component itself. The argument must also consider assumptions made about the system and the operating environment during the development of the ML component that may become invalid during operation. The sufficiency of the identification of these erroneous behaviours must be justified in J6.2. This may be informed by the results of system safety analysis activities.
Claim G6.6 is supported by two sub-claims, one that demonstrates the system design incorporates sufficient monitoring of erroneous behaviours, and one demonstrating that the response of the system to such behaviours is acceptable.

\subsection*{G6.8}
It must be demonstrated that the system design incorporates sufficient monitoring of the identified erroneous behaviour to ensure that any behaviour that could result in violation of a safety requirement will be identified if it occurs during operation.

\subsection*{G6.9}
It must be demonstrated that the system design ensures that an acceptable response can be provided if monitoring reveals erroneous behaviour during operation. The response may take many forms, depending on the nature of the system, the relevant system hazard behaviour and the erroneous behaviour identified. This may include, for example, the provision of redundancy in the system architecture or the specification of safe degraded operation. Evidence should be provided to show that a sufficiently safe response is provided.

\clearpage
\section{Afterword}

It would not have been possible to produce this document without the numerous insightful interactions of the authors with a wide range of experts across industry and academia. We cannot acknowledge them all personally here, but their contributions are very much appreciated. In particular we would like to thank the following AAIP Visiting Fellows and colleagues who kindly reviewed and provided feedback on an initial draft of this document:

\begin{itemize}
    \item Rob Ashmore (DSTL)
    \item Alec Banks (DSTL)
    \item Simon Burton (Fraunhofer IKS)
    \item Jelena Frtunikj (ArgoAI)
    \item Lydia Gauerhof (Robert Bosch GmbH)
    \item Simos Gerasimou (University of York)
    \item Farah Magrabi (Macquarie University)
    \item Mike Parsons (University of York)
    \item Roger Rivett (Jaguar Land Rover (retired))
    \item Simon Smith (CACI)
    \item Mark Sujan (Human Factors Everywhere)
    \item Sean White (NHS Digital)
\end{itemize}

This document provides the first version of the guidance. In the coming months we will undertake validation of the process by applying it to a number of case studies in different domains. The document will be updated to include the details of these case studies.

We would very much value feedback on the guidance provided in this document. We would in particular encourage the reader where appropriate to apply this guidance to the development of systems and share those experiences with the authors.

This work has been funded by Lloyds Register Foundation and the University of York through the Assuring Autonomy International Programme https://www.york.ac.uk/assuring-autonomy.

\clearpage
\bibliographystyle{plainurl}
\bibliography{amlas}

\end{document}